%% file: iclr2027_conference.tex
\documentclass{article} % For LaTeX2e
\usepackage{iclr2027_conference,times}

\input{math_commands.tex}

\usepackage{hyperref}
\usepackage{url}

\usepackage{graphicx}
\usepackage{booktabs}  
\usepackage{arydshln} 
\usepackage{pgf}
\usepackage{xcolor}
\usepackage{colortbl}
\usepackage{pgfmath}
\usepackage{multirow}
\usepackage{subcaption}
\usepackage{cleveref}
\usepackage{titletoc}

\definecolor{reducgreen}{RGB}{95,200,105}
\definecolor{mathbg}{RGB}{231,217,241}
\definecolor{sciencebg}{RGB}{250,225,190}
\definecolor{reducgreen}{RGB}{95,200,105}
\newcommand{\redcell}[1]{%
  \pgfmathtruncatemacro{\shade}{8 + 82*min(#1,80)/80}%
  \edef\applycellcolor{%
    \noexpand\cellcolor{reducgreen!\shade!white}%
  }%
  \applycellcolor
  $-#1\%$%
}

\definecolor{increasered}{RGB}{220,80,80}

\newcommand{\increasecell}[1]{%
  \pgfmathtruncatemacro{\shade}{8 + 82*min(#1,80)/80}%
  \edef\applycellcolor{%
    \noexpand\cellcolor{increasered!\shade!white}%
  }%
  \applycellcolor
  $+#1\%$%
}

\title{MetaCtrl: Your Large Language Models Can Reason Better and More Concisely with a Metacognitive Controller}

\author{
Zhibin Wen\textsuperscript{1},
Tao Han\textsuperscript{2,3},
Lei Bai\textsuperscript{3},
Can Li\textsuperscript{4},
Yang Xu\textsuperscript{1}\thanks{Corresponding author. Correspondence to: \texttt{xuyang@sustech.edu.cn}}
\\[4pt]
\textsuperscript{1}Southern University of Science and Technology \\
\textsuperscript{2}The Hong Kong University of Science and Technology \\
\textsuperscript{3}Shanghai Artificial Intelligence Laboratory \\
\textsuperscript{4}Tongji University
}

\iclrfinalcopy % Uncomment for camera-ready version, but NOT for submission.
\begin{document}

\maketitle

\begin{abstract}

Large reasoning models improve performance on challenging problems by allocating additional computation before answering, but longer reasoning does not always lead to better results and can introduce substantial redundant reasoning on simple problems. Conversely, aggressively shortening reasoning can degrade performance on difficult ones. Effective reasoning therefore requires dynamically deciding when additional computation is useful based on the reasoner’s capabilities and evolving solution state. Existing approaches often rely on predefined budgets or intervention rules, retrain the target reasoner, or require additional supervision.
We introduce \textbf{MetaCtrl}, a lightweight controller that adaptively regulates a frozen reasoner without predefined token budgets or reasoner retraining. We formulate reasoning regulation as a sequential metacognitive control problem: MetaCtrl observes the evolving reasoning trace and decides whether to continue, simplify, skip redundant steps, or conclude reasoning. It is trained directly with reinforcement learning using a reward that prioritizes correctness while favoring shorter trajectories among correct solutions, requiring neither supervised intervention trajectories nor problem-specific budgets.
Across seven benchmarks spanning mathematics, science, and code, MetaCtrl consistently improves the accuracy of LRMs while reducing their reasoning length. On DeepSeek-R1-Distill-Qwen-7B, it improves average accuracy by \textcolor{red}{4.7} points while reducing generation length by \textcolor{red}{53.3\%}. Without further training, the same controller transfers to an unseen reasoner (e.g., Qwen3-14B), improving average accuracy by \textcolor{red}{2.9} points and reducing generation length by \textcolor{red}{50.3\%}. These results establish MetaCtrl as a plug-and-play controller for improving reasoning accuracy while substantially reducing inference-time generation. The code is available at \url{https://github.com/binbin2xs/MetaCtrl}.

\end{abstract}

\begin{figure}[h]
	\centering
	\includegraphics[width=0.99\linewidth]{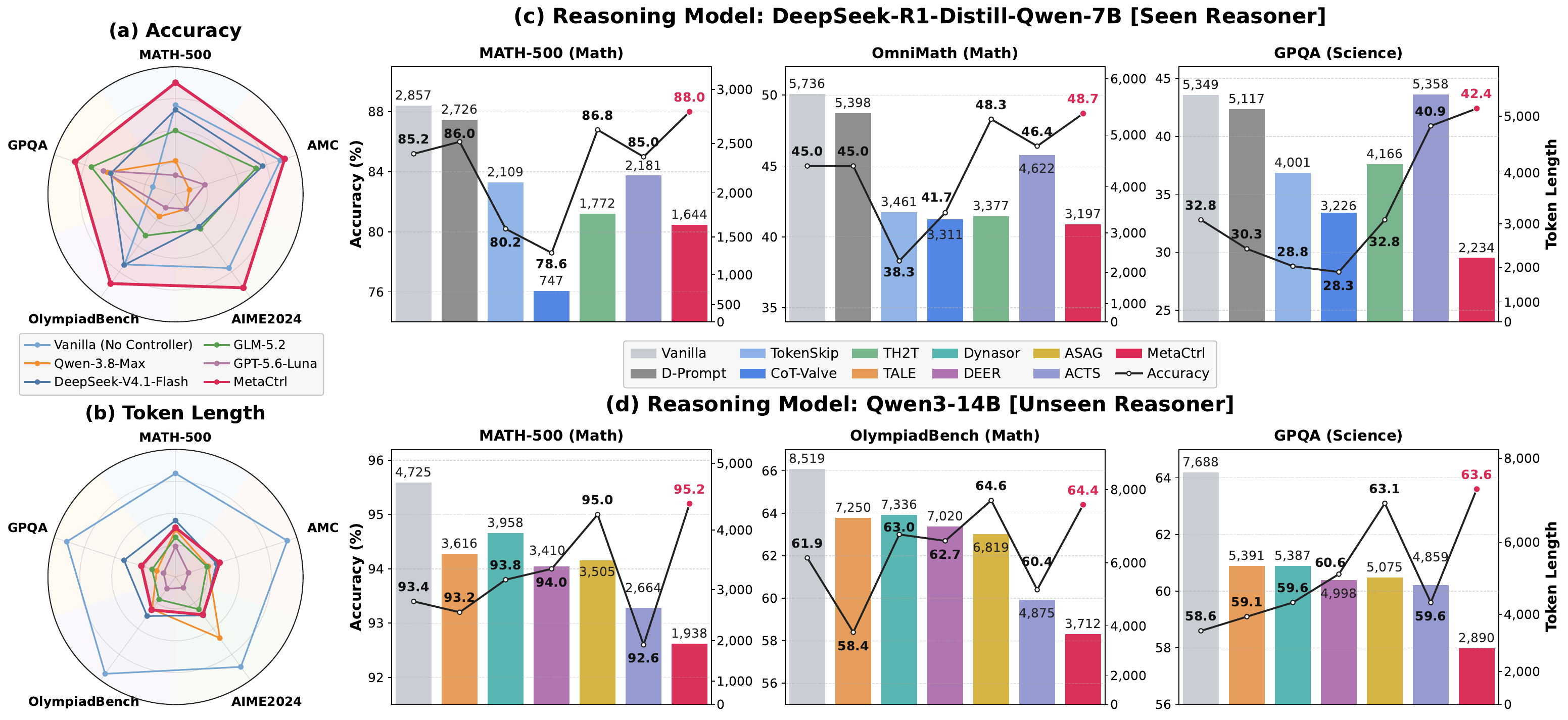}
	\caption{\textbf{Left:} Accuracy and generation length of the same reasoner when controlled by MetaCtrl, controlled by frontier LLMs, or run without external control across five benchmarks. \textbf{Right:} Comparison of MetaCtrl with representative efficient reasoning baselines across two reasoning models.}
    \label{fig:intro}
\end{figure}

\section{Introduction}

Recent large reasoning models (LRMs) \citep{jaech2024openai, guo2025deepseek, yang2025qwen3} have shown that scaling test-time computation \citep{snell2025scaling} through chain of thought \citep{kojima2022large,wei2022chain,yao2023tree,sprague2025cot} reasoning can substantially improve performance on challenging tasks \citep{phan2025humanity,li2025s,cao2025step,yu2025chain,chen2026reasoning}. However, longer reasoning steps do no always lead to better performance. LRMs often spend excessive computation on redundant verification and unnecessary exploration, and may even deviate from an initially correct solution after prolonged deliberation \citep{chen2025not,zhou2026more,zhang2026llms,li2026stop}. Conversely, overly restricting reasoning can hurt difficult problems that genuinely require more computation \citep{madaan2023self, jin2024impact}. The key challenge is therefore not simply to make reasoning shorter or longer, but to determine whether additional reasoning remains beneficial and to allocate test-time computation accordingly.

This challenge has a close analogue to the classical concept of ``metacognition'' in cognitive science, which is commonly characterized as the monitoring and control of one’s own cognitive processes \citep{ackerman2017meta}. The Nelson–Narens framework \citep{nelson1990metamemory} characterizes metacognition as the interaction between a meta-level and an object-level processes, with the former monitoring and steering the state of the latter. 
Applying this concept to reasoning LLMs, it is natural to propose the idea of \emph{metareasoning} strategy that regulates cognitive effort and strategy, including whether to continue, redirect, or terminate a reasoning process \citep{lieder2017strategy}. 
% Because human cognition is inherently bounded, resource-rational accounts further suggest that cognitive effort should be allocated according to the expected benefits of additional computation relative to its costs, rather than expended uniformly \citep{lieder2020resource,callaway2022rational}. Effective reasoning therefore requires not only solving the problem, but also regulating when further computation is worth pursuing.

We posit that overthinking in LRMs can be understood as a failure of resource-rational metareasoning. At any point in a reasoning trajectory, additional computation is worthwhile only when its expected benefit outweighs its cost. Continuing after this marginal utility has vanished leads to overthinking, whereas terminating while further computation is still valuable leads to underthinking. Crucially, the computational demand of a problem is not determined by the input alone\citep{liu2026think,jia2026makes}: it depends on the reasoner's capability and evolves dynamically as the reasoning trajectory unfolds. For example, the same problem may be trivial for a stronger reasoner yet require substantial deliberation for a weaker one. Moreover, the same problem may require substantial computation early on, yet little or none once a promising solution path or sufficient solution has been reached. Therefore, efficient reasoning should be viewed as a dynamic resource-allocation problem, where the value of further reasoning must be continually reassessed as the process unfolds.
% sequential computation-allocation problem

Existing approaches only partially address this dynamic allocation problem. Prompt-based methods \citep{aytes2025sketch,ding2024break,han2025token,xu2025chain} estimate problem difficulty from the input and predict a reasoning budget/strategy before generation. However, the input alone provides only limited evidence about how much computation a particular reasoner will need, making the accurate allocation difficult. Model-based methods \citep{liu2026think,yang2025towards,ma2025cot,xia2025tokenskip} use supervised fine-tuning (SFT) or reinforcement learning (RL) to induce concise or adaptive reasoning, but they entangle object-level reasoning with meta-level regulation, which may cause disturbance to the reasoner. % typically require additional data construction and post-training of the target reasoner. 
Output-based methods \citep{li2026stop,wang2025sampling,yang2026dynamic,huang2026efficient} instead regulate computation during inference through compression, pruning, early termination, or other interventions, but often rely on predefined proxy signals or decision rules whose reliability may vary across scenarios. 
While effective, these approaches do not fully address the trajectory-level challenge of regulating computation online, without retraining the reasoner or relying on predefined allocation rules.

This gap naturally motivates a resource-rational metacognitive approach to reasoning control, where computation is allocated online according to the problem and the evolving reasoning trajectory rather than prescribed in advance or encoded into the reasoner itself. We therefore formulate efficient reasoning as a sequential metacognitive control problem and introduce \textbf{MetaCtrl}, a learned meta-level controller that regulates a frozen object-level reasoner. The controller repeatedly observes the evolving reasoning state and dynamically determines how reasoning should proceed through four lightweight interventions: \textit{continue} leaves reasoning unchanged, \textit{fast-think} encourages concise derivation, \textit{skip-think} bypasses redundant intermediate reasoning, and \textit{stop-think} triggers a bounded conclusion before terminating deliberation, while preserving the reasoning already generated. Rather than relying on supervised intervention trajectories, fixed human-specified reasoning budgets, or hand-crafted rules, we train the controller directly with Group Relative Policy Optimization (GRPO) \citep{shao2024deepseekmath} to learn when and how to regulate reasoning from trajectory-level outcomes. The reward signal jointly reflects final correctness and reasoning efficiency, providing credit to the sequence of meta-level decisions according to their eventual effect on task success and computation cost.

We evaluate \textbf{MetaCtrl} on seven reasoning benchmarks spanning mathematics, science, and code. On the seen DeepSeek-R1-Distill-Qwen-7B reasoner, MetaCtrl improves accuracy while reducing generation length by \textbf{47.3\%}, \textbf{58.2\%}, and \textbf{72.1\%} on mathematical, scientific, and code reasoning, respectively. More importantly, the controller transfers to multiple unseen reasoners without model-specific training. On Qwen3-14B, it reduces generation length by \textbf{51.6\%}, \textbf{62.4\%}, and \textbf{33.2\%} across the three domains, respectively, while improving accuracy across all evaluated benchmarks. Notably, compared with DeepSeek-V4.1-Flash (552B parameters) \citep{xu2026deepseek} as the controller under the same frozen reasoner, MetaCtrl (4B parameters) improves average accuracy by 7.6 percentage points while reducing average generation length by 7.3\% across the five benchmarks. Overall, our contributions can be outlined as follows:

\textbf{(1) Resource-rational metacognitive formulation.}
We frame overthinking and underthinking in LRMs as failures to appropriately regulate inference-time computation, and introduce \textbf{MetaCtrl}, a resource-rational metacognitive framework that dynamically regulates reasoning effort according to the reasoner and its evolving reasoning trajectory.

\textbf{(2) Learning metacognitive control from outcomes.}
To learn when and how to regulate reasoning, we introduce a metacognitive policy learning approach that uses GRPO to optimize the sequential interventions in a frozen reasoner. Final correctness and reasoning efficiency jointly guide policy learning, eliminating the need for supervised intervention trajectories.

\textbf{(3) Effective and transferable reasoning regulation.}
% Comprehensive experiments show that MetaCtrl consistently improves reasoning accuracy and efficiency over strong baselines, and transfer well across different reasoning models without additional training.
Comprehensive experiments show that MetaCtrl consistently improves the reasoning accuracy and efficiency of LRMs over strong baselines, and generalizes well to unseen reasoning models without additional training.

\section{Related Work}

\begin{figure}[t]
	\centering
	\includegraphics[width=\linewidth]{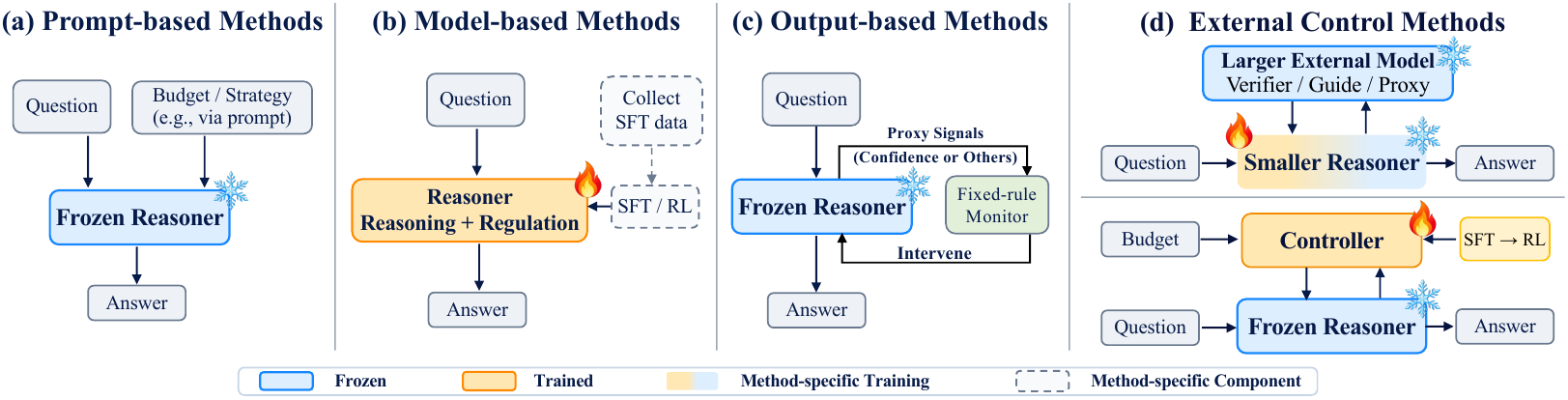}
	\caption{Comparison of four paradigms for efficient LRMs inference. In external control methods, some approaches train the reasoner, while others keep it frozen.}
    \label{fig:related_work}
\end{figure}

Following \citet{sui2025stop}, we group efficient reasoning methods into three categories and additionally discuss external reasoning control methods that are most closely related to ours, as illustrated in Figure~\ref{fig:related_work}. \textbf{Prompt-based methods} improve efficiency without modifying reasoner parameters \citep{lee2025well,yu2025premise,zhang2025long,renze2024benefits,xu2025chain,han2025token,aytes2025sketch,liang2025thinkswitcher,liu2026diffadapt}. They shorten reasoning or select budgets and reasoning strategies from the input. However, it is difficult to determine from the question alone how much and how to reason, and decisions made before generation cannot adapt to subsequent changes in reasoning progress. MetaCtrl instead regulates computation online by conditioning interventions on the evolving reasoning trajectory. \textbf{Model-based methods} internalize efficient reasoning through post-training \citep{kang2025c3ot,ma2025cot,yang2025think,yu2024distilling,xia2025tokenskip,luo2026o1,hou2025thinkprune,aggarwal2025l1,shen2025dast,liu2026think}. These approaches train the reasoner itself to shorten, compress, or adapt its reasoning, often through SFT, RL, or explicit length constraints. In contrast, MetaCtrl freezes the reasoner and decouples problem solving from computation control, enabling the same controller to transfer across reasoners without retraining them. This separation becomes increasingly advantageous as reasoner scale grows, since only the lightweight controller requires training. \textbf{Output-based methods} regulate ongoing generation using signals such as confidence, certainty, hidden states, attention, or logits \citep{liao2025reward,liu2025answer,tikhonov2026confidence,hao2024training,yang2026dynamic,fu2025efficiently,chen2025seal,li2026stop}. Many rely on predefined decision rules or thresholds, or require access to internal model signals, making their behavior sensitive to the chosen control criteria. MetaCtrl instead learns trajectory-conditioned interventions directly from correctness--efficiency rewards using only the observable reasoning trace. \textbf{External reasoning control} uses an auxiliary model to regulate a separate reasoner \citep{pan2025specreason,yang2025speculative,wang2026intervene,xia2026agentic}. Prior methods may rely on substantially larger external models or even human feedback, and often participate directly in problem solving by generating, verifying, correcting, or providing solution-specific guidance, effectively coupling reasoning across multiple models. Others additionally retrain the reasoner or require distilled supervision and predefined token budgets. MetaCtrl instead learns a lightweight, budget-free controller through reinforcement learning, using only task-agnostic control actions to regulate a frozen reasoner without supplying solution content or requiring supervised intervention trajectories. Appendix~\ref{app:related_work} provides a more detailed discussion.

\begin{figure}[t]
	\centering
	\includegraphics[width=\linewidth]{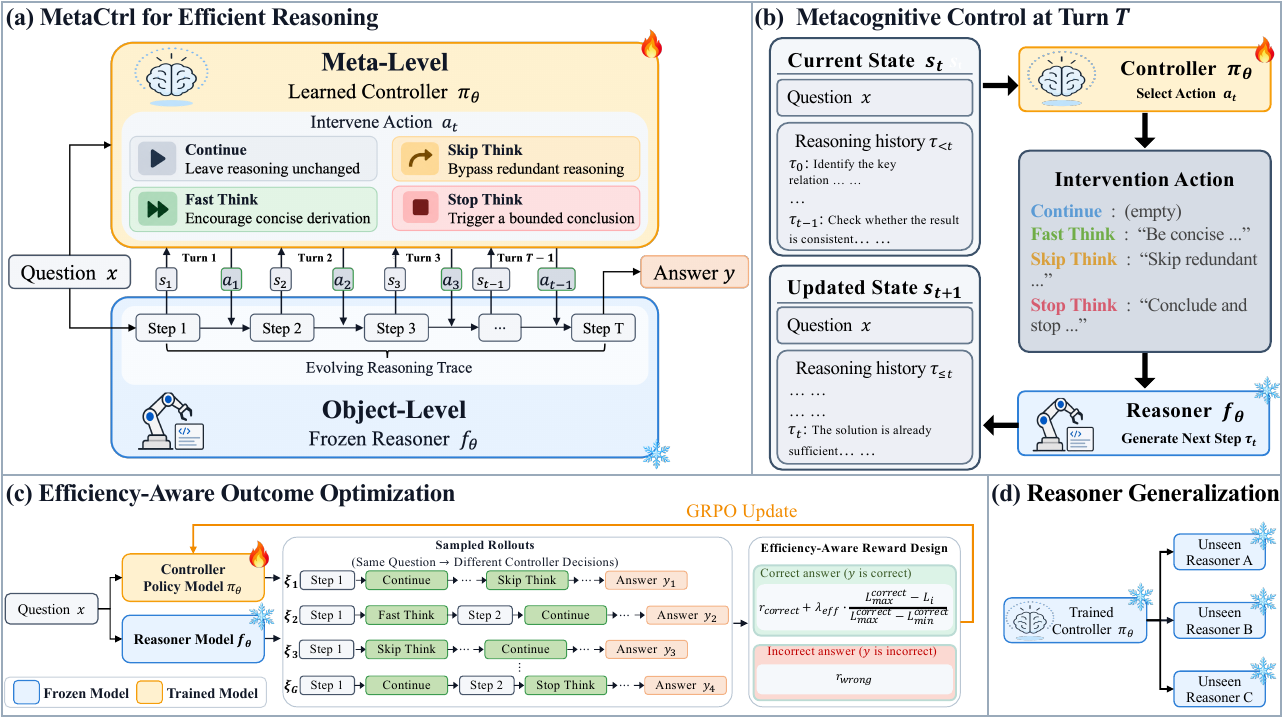}
	\caption{\textbf{Overview of MetaCtrl.} (a) A learned meta-level controller regulates a frozen reasoner through four intervention actions, without an input budget. (b) At each turn, it selects an intervention based on the current reasoning state. (c) The controller is trained with GRPO using an efficiency-aware outcome reward. (d) The trained controller generalizes to unseen reasoners without retraining.}
    \label{fig:framework}
\end{figure}

\section{MetaCtrl: Metacognitive Reasoning Control}

% We aim to bridge the gap between reasoning methods that require model adaptation and inference-time control with predefined strategies or budgets. We present \textbf{MetaCtrl}, a metacognitive control framework that dynamically regulates a frozen reasoner from its evolving reasoning trajectory, without modifying the reasoner or specifying its computation budget in advance.

\subsection{Generation with Metacognitive Control}

% Viewing overthinking as a failure of resource-rational metareasoning
% We formulate efficient reasoning as a \emph{sequential computation-allocation problem}. 
We formulate reasoning as a generation process controlled by metacognitive actions. 
Let $x$ denote an input problem, $f_{\phi}$ a frozen object-level reasoner, and $\pi_{\theta}$ a trainable meta-level controller. 
% Rather than prescribing the amount of computation from the input alone, MetaCtrl regulates the reasoning process based on its evolving trajectory. 
We assume that the final successful reasoning trajectory can be decomposed into multiple sub-traces:
\[
[\tau_0,a_1,\tau_1,\dots,a_T,\tau_T],
\]
in which $a_t$ is an \textbf{action} sampled from the controller's discrete output space $\mathcal{A}$ (\cref{eq:action_space}), determined by all preceding traces and actions: 
% At control step $t$, the controller observes the current reasoning state
% \[
% s_t=(x,\tau_t),
% \]
% where $\tau_t$ contains the current reasoning trace up to control step $t$ and all previous control decisions. 
% Based on $s_t$ a new control action is selected: 
\[
a_{t} \sim \pi_{\theta}(\cdot\mid x \oplus \tau_0 \oplus a_1 \cdots \oplus a_{t-1} \oplus \tau_{t-1}),
\]
% The action $a_{t+1}$ is selected from a discrete space $\mathcal{A}$ (\cref{eq:action_space}), which determines how the frozen reasoner should proceed to the next state. 
where $\oplus$ denotes concatenation of traces and actions with the relative order preserved. Let $s_t$ denote the condition as a state, then the action selection  essentially from a Markovian policy $\pi_\theta(\cdot\mid s_t)$. The newly selected action $a_t$ produces a new trace $\tau_{t}$ from the reasoner $f_\phi$. This process repeats for $T$ steps and generate a full controlled reasoning trajectory $\xi=(\tau_0,a_1,\tau_1,\ldots,a_T,\tau_T)$, which terminates with a final answer $y_{\xi}$. Therefore, given a dataset $\mathcal{D}$, we optimize the controller $\theta$ to balance reasoning performance and efficiency:
\[
\max_{\theta}\;
\mathbb{E}_{(x,y^*)\sim\mathcal D,\,
\xi\sim p_{\theta,\phi}(\cdot\mid x)}
\left[R(\xi,y^*)\right],
\]
where $y^*$ denotes the ground-truth answer and $R$ jointly evaluates answer correctness and object-level reasoning cost. The reasoner parameters $\phi$ remain frozen throughout training.

\subsection{Action Space of Metacognitive Control}

The action space of metacognitive controls $a_t$ matter to the final performance. MetaCtrl's basic principle is to place a lightweight text-modality interventions on the evolving reasoning trace. 
% We regulate the reasoner through lightweight textual interventions on its evolving reasoning trace, without updating its parameters or accessing its internal states, because we aim for a model-agnostic control interface that can transfer across different frozen reasoners.
% \textbf{Action space design.}
Concretely, at each control point, we select from a discrete action space
\[
\mathcal A=
\{
\textsc{Continue},
\textsc{Fast Think},
\textsc{Skip Think},
\textsc{Stop Think}
\}.\label{eq:action_space}
\]
\textsc{Continue} preserves the context and lets the reasoner proceed naturally. \textsc{Fast Think} encourages a more concise continuation, while \textsc{Skip Think} more aggressively directs the reasoner toward the solution with only the necessary intermediate reasoning. \textsc{Stop Think} signals that the current reasoning is sufficient and prompts the reasoner to conclude and finalize its answer. Except for \textsc{Continue}, each action is implemented by appending a short textual prompt to the current context. The exact prompts are provided in Appendix~\ref{app:intervention_prompts}.

\textbf{Where to intervene.}
Control is exposed only at natural language boundaries. After the initial uncontrolled trace $\tau_0$, we let generation pause at sentence-ending paragraph boundaries, implemented using \texttt{".\textbackslash n\textbackslash n"}, \texttt{"?\textbackslash n\textbackslash n"}, and two whitespace variants \texttt{". \textbackslash n\textbackslash n"}, \texttt{"? \textbackslash n\textbackslash n"}. The controller is invoked between coherent reasoning traces rather than at arbitrary token positions, providing semantically meaningful states while leaving token-level generation \emph{uninterrupted}. 
% If the reasoner emits \texttt{</think>}, reasoning terminates naturally and no further intervention is applied.

% \textbf{When to intervene.}
% The action is sampled from the policy distribution $\pi_{\theta}(\cdot\mid s_t)$ conditioned on the evolving interaction, rather than determined solely from the problem input in advance or by a fixed stopping rule, confidence threshold, or problem-specific reasoning budget. 
% The controller can therefore adapt computation online as reasoning progresses: continuing when the current reasoning remains incomplete, compressing redundant deliberation, or concluding when further reasoning provides little benefit.

\textbf{When to intervene.} The action is sampled from the policy distribution $\pi_{\theta}(\cdot\mid s_t)$ conditioned on the evolving interaction. In contrast, problem-level allocation methods select a reasoning strategy or budget before generation according to $\pi(\cdot\mid x)$, while output-based methods often trigger early exit through predefined proxy criteria, e.g., stopping when the confidence $C_t$ of an induced trial answer exceeds a fixed threshold $\delta$. Conditioning on $s_t$ allows MetaCtrl to determine when to intervene from the semantic content of the reasoning trace, enabling flexible control as reasoning evolves.

\subsection{Outcome-Guided Policy Optimization for Efficient Metareasoning}

We train MetaCtrl directly from the outcomes of its interactions with the frozen reasoner, without expert intervention trajectories or intermediate action labels. For each problem $x$, we sample a group of $G$ controlled reasoning trajectories $\{\xi_i\}_{i=1}^{G}$ using the controller $\pi_\theta$, and evaluate each trajectory by both answer correctness and reasoner generation length.

\textbf{Correctness-gated efficiency reward.}
Let $c_i\in\{0,1\}$ indicate whether $\xi_i$ produces the correct answer, and let $L_i$ be the total number of tokens generated by the object-level reasoner, including both reasoning and final answer generation.

To prevent efficiency optimization from favoring prematurely terminated but incorrect reasoning, we reward shorter computation \emph{only among correct trajectories}. Define
\[
\mathcal I^{\mathrm{correct}}=\{i\mid c_i=1\},
\qquad
L_{\min}^{\mathrm{correct}}=\min_{i\in\mathcal I^{\mathrm{correct}}}L_i,
\qquad
L_{\max}^{\mathrm{correct}}=\max_{i\in\mathcal I^{\mathrm{correct}}}L_i.
\]
For a correct trajectory, its relative efficiency is
\[
e_i=
\frac{L_{\max}^{\mathrm{correct}}-L_i}
     {L_{\max}^{\mathrm{correct}}-L_{\min}^{\mathrm{correct}}},
\]
with $e_i=0$ when all correct trajectories have the same length. The reward is
\[
R_i=
\begin{cases}
r_{\mathrm{correct}}
+\lambda_{\mathrm{eff}}e_i,
& c_i=1,\\[2mm]
r_{\mathrm{wrong}},
& c_i=0,
\end{cases}
\]
where $r_{\mathrm{correct}}>0$ is the base correctness reward, $\lambda_{\mathrm{eff}}>0$ controls the strength of the efficiency bonus, and $r_{\mathrm{wrong}}<0$ is the penalty for an incorrect trajectory. Thus, shorter trajectories are favored only among correct solutions, while all incorrect trajectories receive the same length-independent penalty, encouraging efficient reasoning without trading correctness for premature termination.

\textbf{Online GRPO update.}
We optimize the controller online with GRPO
\citep{shao2024deepseekmath} over trajectories sampled for the same problem. Following the group-normalization treatment of Dr.~GRPO
\citep{liu2025understanding}, we omit within-group standard-deviation normalization and use the mean-centered advantage
\[
\hat A_i
=
R_i-\bar R,
\qquad
\bar R
=
\frac{1}{G}\sum_{j=1}^{G}R_j.
\]
The trajectory-level advantage $\hat A_i$ is assigned to the controller generated action tokens in $\xi_i$, while reasoner generated tokens are masked from the policy loss. Consequently, policy optimization updates only $\pi_{\theta}$, with $f_{\phi}$ remaining fixed.

Through outcome-guided optimization, MetaCtrl learns trajectory-dependent interventions that improve both reasoning correctness and efficiency, without expert intervention trajectories or intermediate action labels. Rather than imitating a fixed reasoning schedule, the controller learns to allocate computation dynamically from the relative outcomes of its own interactions with the frozen reasoner.

\begin{table*}[t]
\caption{Comparison across reasoning models of different scales and and multiple baselines.}
\label{tab:dsr_main_results}
\centering
\small
\setlength{\tabcolsep}{3.8pt}
\renewcommand{\arraystretch}{1.12}

\resizebox{\textwidth}{!}{
\begin{tabular}{lccc ccc ccc ccc |cc}
\toprule
& \multicolumn{9}{c}{\cellcolor{mathbg}\textbf{Math Reasoning}}
& \multicolumn{3}{c}{\cellcolor{sciencebg}\textbf{Scientific Reasoning}}
& \multicolumn{2}{c}{} \\
\cmidrule(lr){2-10}
\cmidrule(lr){11-13}
\cmidrule(lr){14-15}

\textbf{Methods}
& \multicolumn{3}{c}{\textbf{MATH-500}}
& \multicolumn{3}{c}{\textbf{AIME2024}}
& \multicolumn{3}{c}{\textbf{OmniMath}}
& \multicolumn{3}{c}{\textbf{GPQA Diamond}}
& \multicolumn{2}{|c}{\textbf{AVG}} \\
\cmidrule(lr){2-4}
\cmidrule(lr){5-7}
\cmidrule(lr){8-10}
\cmidrule(lr){11-13}
\cmidrule(lr){14-15}

& \textbf{Acc.} & \textbf{Len.} & \textbf{Reduc.}
& \textbf{Acc.} & \textbf{Len.} & \textbf{Reduc.}
& \textbf{Acc.} & \textbf{Len.} & \textbf{Reduc.}
& \textbf{Acc.} & \textbf{Len.} & \textbf{Reduc.}
& \textbf{Acc.$\uparrow$} & \textbf{Reduc.$\uparrow$} \\
\midrule

% =========================================================
% DeepSeek-R1-Distill-Qwen-7B
% =========================================================

\multicolumn{15}{c}{
    \textit{\textbf{DeepSeek-R1-Distill-Qwen-7B}} \textbf{[Seen Reasoner]}
} \\
\midrule

NoThinking
& 79.4 & {699} & -75.5
& 40.0 & {4134} & -60.9
& 40.0 & {2083} & -63.7
& 29.3 & {2084} & -61.0
& 47.2 & -65.3 \\
\midrule

Vanilla
& 85.2 & 2857 & \redcell{0}
& {50.0} & 10570 & \redcell{0}
& 45.0 & 5736 & \redcell{0}
& 32.8 & 5349 & \redcell{0}
& 53.3 & \redcell{0} \\

D-Prompt
& 86.0 & 2726 & \redcell{4.6}
& 46.7 & 10209 & \redcell{3.4}
& 45.0 & 5398 & \redcell{5.9}
& 30.3 & 5117 & \redcell{4.3}
& 52.0 & \redcell{4.6} \\

TokenSkip
& 80.2 & 2109 & \redcell{26.2}
& 40.0 & {5559} & \redcell{47.4}
& 38.3 & 3461 & \redcell{39.7}
& 28.8 & 4001 & \redcell{25.2}
& 46.8 & \redcell{34.6} \\

CoT-Valve
& 78.6 & \textbf{{747}} & \redcell{73.9}
& 43.3 & \textbf{5871} & \redcell{44.5}
& 41.7 & 3311 & \redcell{42.3}
& 28.3 & 3226 & \redcell{39.7}
& 48.0 & \redcell{50.1} \\

AdaCtrl
& 74.0 & 3196 & \increasecell{11.9}
& 21.3 & 16889 & \increasecell{59.8}
& - & - & -
& - & - & -
& - & - \\

TH2T
& {86.8} & 1772 & \redcell{38.0}
& {50.0} & 7490 & \redcell{29.1}
& {48.3} & 3377 & \redcell{41.1}
& 32.8 & 4166 & \redcell{22.1}
& {54.5} & \redcell{32.6} \\

ACTS
& 85.0 & 2181 & \redcell{23.7}
& 36.7 & 5909 & \redcell{44.1}
& 46.4 & 4622 & \redcell{19.4}
& {40.9} & 5358 & \increasecell{0.2}
& 52.3 & \redcell{21.8} \\

\rowcolor{gray!18}
\textbf{MetaCtrl}
& \textbf{88.0} & 1644 & \redcell{42.5}
& \textbf{56.7} & 5899 & \redcell{44.2}
& \textbf{48.7} & \textbf{3197} & \redcell{44.3}
& \textbf{42.4} & \textbf{2234} & \redcell{58.2}
& \textbf{59.0} & \redcell{47.3} \\

% =========================================================
% DeepSeek-R1-Distill-Qwen-32B
% =========================================================

\midrule
\multicolumn{15}{c}{
    \textit{\textbf{DeepSeek-R1-Distill-Qwen-32B}} \textbf{[Unseen Reasoner]}
} \\
\midrule

NoThinking
& 80.2 & {659} & -72.0
& 53.3 & {3208} & -66.6
& 45.0 & {2146} & -61.3
& 50.5 & {1627} & -62.9
& 57.3 & -65.7 \\
\midrule

Vanilla
& {87.2} & 2357 & \redcell{0}
& \textbf{60.0} & 9605 & \redcell{0}
& 50.0 & 5540 & \redcell{0}
& 52.0 & 4384 & \redcell{0}
& \textbf{62.3} & \redcell{0} \\

D-Prompt
& 86.8 & 2198 & \redcell{6.8}
& {56.7} & 9445 & \redcell{1.7}
& 48.3 & 5135 & \redcell{7.3}
& 49.5 & 3877 & \redcell{11.6}
& 60.3 & \redcell{6.9} \\

TokenSkip
& 79.8 & 1567 & \redcell{33.5}
& 50.0 & 5856 & \redcell{39.0}
& 43.3 & 3730 & \redcell{32.7}
& 47.4 & 2874 & \redcell{34.4}
& 55.1 & \redcell{34.9} \\

CoT-Valve
& 85.3 & 2263 & \redcell{3.9}
& 53.3 & 7997 & \redcell{16.7}
& 43.3 & 4663 & \redcell{15.8}
& 49.0 & 3550 & \redcell{19.0}
& 57.7 & \redcell{13.9} \\

TH2T
& \textbf{88.0} & 1733 & \redcell{26.5}
& {56.7} & 7004 & \redcell{27.1}
& 48.3 & \textbf{2998} & \redcell{45.9}
& 52.0 & 2367 & \redcell{46.0}
& 61.3 & \redcell{36.4} \\

ACTS
& 85.0 & 2132 & \redcell{9.5}
& 43.3 & 6323 & \redcell{34.2}
& {50.3} & 4539 & \redcell{18.1}
& {52.5} & 5036 & \increasecell{14.9}
& 57.8 & \redcell{11.7} \\

\rowcolor{gray!18}
\textbf{MetaCtrl}
& \textbf{88.0} & \textbf{1528} & \redcell{35.2}
& \underline{56.7} & \textbf{4951} & \redcell{48.5}
& \textbf{50.6} & 3113 & \redcell{43.8}
& \textbf{53.2} & \textbf{1968} & \redcell{55.1}
& {62.1} & \redcell{45.6} \\

\bottomrule
\end{tabular}
}
\vspace{-0.25cm}
\end{table*}

\begin{table*}[t]
\caption{Comparison across the Qwen3 series at different model scales and multiple baselines.}
\label{tab:qwen3_main_results}
\centering
\small
\setlength{\tabcolsep}{3.8pt}
\renewcommand{\arraystretch}{1.12}

\resizebox{\textwidth}{!}{
\begin{tabular}{lccc ccc ccc ccc |cc}
\toprule

& \multicolumn{9}{c}{\cellcolor{mathbg}\textbf{Math Reasoning}}
& \multicolumn{3}{c}{\cellcolor{sciencebg}\textbf{Scientific Reasoning}}
& \multicolumn{2}{c}{} \\
\cmidrule(lr){2-10}
\cmidrule(lr){11-13}
\cmidrule(lr){14-15}

\textbf{Methods}
& \multicolumn{3}{c}{\textbf{MATH-500}}
& \multicolumn{3}{c}{\textbf{AIME 2024}}
& \multicolumn{3}{c}{\textbf{OlympiadBench}}
& \multicolumn{3}{c}{\textbf{GPQA Diamond}}
& \multicolumn{2}{|c}{\textbf{AVG}} \\
\cmidrule(lr){2-4}
\cmidrule(lr){5-7}
\cmidrule(lr){8-10}
\cmidrule(lr){11-13}
\cmidrule(lr){14-15}

& \textbf{Acc.} & \textbf{Len.} & \textbf{Reduc.}
& \textbf{Acc.} & \textbf{Len.} & \textbf{Reduc.}
& \textbf{Acc.} & \textbf{Len.} & \textbf{Reduc.}
& \textbf{Acc.} & \textbf{Len.} & \textbf{Reduc.}
& \textbf{Acc.$\uparrow$} & \textbf{Reduc.$\uparrow$} \\
\midrule

% =========================================================
% Qwen3-8B
% =========================================================

\multicolumn{15}{c}{
    \textit{\textbf{Qwen3-8B}} \textbf{[Unseen Reasoner]}
} \\
\midrule
NoThinking
& 87.4 & 1480 & -70.0
& 23.3 & 7121 & -41.2
& 48.7 & 5219 & -43.7
& 48.5 & 2564 & -72.7
& 52.0 & -56.9 \\
\midrule
Vanilla
& 92.2 & 4926 & \redcell{0}
& {63.3} & 12101 &  \redcell{0}
& 59.9 & 9268 &  \redcell{0}
& 52.5 & 9382 &  \redcell{0}
& 67.0 &  \redcell{0} \\

TALE
& 91.8 & 3682 & \redcell{25.3}
& 60.0 & 11847 & \redcell{2.1}
& 56.1 & 7306 & \redcell{21.2}
& 52.0 & 4161 & \redcell{55.6}
& 65.0 & \redcell{26.0} \\

Dynasor
& 92.4 & 3361 & \redcell{31.8}
& 60.0 & 11365 & \redcell{6.1}
& 62.4 & 7640 & \redcell{17.6}
& 55.1 & 5647 & \redcell{39.8}
& 67.5 & \redcell{23.8} \\

DEER
& 92.4 & 2966 & \redcell{39.8}
& 63.3 & 8930 & \redcell{26.2}
& 61.7 & 7367 & \redcell{20.5}
& 54.5 & 5334 & \redcell{43.1}
& 68.0 & \redcell{32.4} \\

ASAG
& 93.0 & 3152 & \redcell{36.0}
& \textbf{66.7} & 8683 & \redcell{28.3}
& 63.9 & 7296 & \redcell{21.3}
& 56.1 & 5714 & \redcell{39.1}
& 69.9& \redcell{31.2} \\

ACTS
& 90.2 & 2736 & \redcell{44.5}
& 46.7 & \textbf{7148}& \redcell{40.9}
& 58.0 & 5059& \redcell{45.4}
& \textbf{57.6} & 5102 & \redcell{45.6}
& 63.1 & \redcell{44.1} \\

\rowcolor{gray!18}
\textbf{MetaCtrl}
& \textbf{94.0} & \textbf{2294} & \redcell{53.4}
& \textbf{66.7} & 8645 & \redcell{28.6}
& \textbf{64.0} & \textbf{4134} & \redcell{55.4}
& {57.1} & \textbf{3518} & \redcell{62.5}
& \textbf{70.5} & \redcell{50.0} \\

% =========================================================
% Qwen3-14B
% =========================================================

\midrule
\multicolumn{15}{c}{
    \textit{\textbf{Qwen3-14B}} \textbf{[Unseen Reasoner]}
} \\
\midrule
NoThinking
& 88.0 & {1403} & -70.3
& 30.0 & 7726 & -26.7
& 50.4 & 5462 & -35.9
& 50.5 & {2308} & -70.0
& 54.7 & -50.7\\
\midrule
Vanilla
& 93.4 & 4725 & \redcell{0}
& {70.0} & 10537 & \redcell{0}
& 61.9 & 8519 & \redcell{0}
& 58.6 & 7688 & \redcell{0}
& 71.0 & \redcell{0}\\

TALE
& 93.2 & 3616 & \redcell{23.5}
& {70.0} & 10358 & \redcell{1.7}
& 58.4 & 7250 & \redcell{14.9}
& 59.1 & 5391 & \redcell{29.9}
& 70.2 & \redcell{17.5} \\

Dynasor
& 93.8 & 3958 & \redcell{16.2}
& {70.0} & 11156 & \increasecell{5.9}
& 63.0 & 7336 & \redcell{13.9}
& 59.6 & 5387 & \redcell{29.9}
& 71.6 & \redcell{13.5} \\

DEER
& 94.0 & 3410 & \redcell{27.8}
& \textbf{73.3} & 8402 & \redcell{20.3}
& 62.7 & 7020 & \redcell{17.6}
& 60.6 & 4998 & \redcell{35.0}
& 72.7 & \redcell{25.2} \\

ASAG
& {95.0} & 3505 & \redcell{25.8}
& \textbf{73.3} & 7602 & \redcell{27.9}
& \textbf{64.6} & 6819 & \redcell{20.0}
& {63.1} & 5075 & \redcell{34.0}
& {74.0} & \redcell{26.9} \\

ACTS
& 92.6 & 2664 & \redcell{43.6}
& 53.3 & {6861} & \redcell{34.9}
& 60.4 & {4875} & \redcell{42.8}
& 59.6 & 4859 & \redcell{36.8}
& 66.5 & \redcell{39.5} \\

\rowcolor{gray!18}
\textbf{MetaCtrl}
& \textbf{95.2} & \textbf{1938} & \redcell{59.0}
& \textbf{73.3} & \textbf{6627} & \redcell{37.1}
& {64.4} & \textbf{3712} & \redcell{56.4}
& \textbf{63.6} & \textbf{2890} & \redcell{62.4}
& \textbf{74.1} & \redcell{53.7} \\

\bottomrule
\end{tabular}
}
\end{table*}

\section{Experiments}

\subsection{Experimental Setup}
We evaluate our method on seven benchmarks spanning mathematical reasoning, scientific reasoning, and code reasoning, using multiple target reasoners and comparing against a broad set of recent and strong baselines. Full details of the training data, evaluation benchmarks and metrics, backbone models, baselines and implementation settings are provided in Appendix~\ref{app:experimental_details}, while prompting details are provided in Appendix~\ref{app:prompt_template}.

\begin{table*}[t]
\caption{Baseline comparison with frontier LLMs as controllers. The reasoning model is DeepSeek-R1-Distill-Qwen-7B.}
\label{tab:api_controller_baselines}
\centering
\small
\setlength{\tabcolsep}{4.0pt}
\renewcommand{\arraystretch}{1.12}

\resizebox{\textwidth}{!}{
\begin{tabular}{l |ccc ccc ccc ccc ccc}
\toprule

\multirow{2}{*}{\textbf{Controller Model}}
& \multicolumn{3}{c}{\textbf{MATH-500}}
& \multicolumn{3}{c}{\textbf{AMC}}
& \multicolumn{3}{c}{\textbf{AIME2024}}
& \multicolumn{3}{c}{\textbf{OlympiadBench}}
& \multicolumn{3}{c}{\textbf{GPQA Diamond}} \\
\cmidrule(lr){2-4}
\cmidrule(lr){5-7}
\cmidrule(lr){8-10}
\cmidrule(lr){11-13}
\cmidrule(lr){14-16}

& \textbf{Acc.} & \textbf{Len.} & \textbf{Reduc.}
& \textbf{Acc.} & \textbf{Len.} & \textbf{Reduc.}
& \textbf{Acc.} & \textbf{Len.} & \textbf{Reduc.}
& \textbf{Acc.} & \textbf{Len.} & \textbf{Reduc.}
& \textbf{Acc.} & \textbf{Len.} & \textbf{Reduc.} \\
\midrule

Vanilla (No Controller)
& 85.2 & 2857 & 
& 73.5 & 7279 & 
& 50.0 & 10570 & 
& 48.6 & 8347 & 
& 32.8 & 5349 &  \\

\midrule

Qwen-3.8-Flash
& 71.4 & 6548 & +129.2\%
& 60.2 & 9229 & +26.8\%
& -- & -- & --
& -- & -- & --
& 30.3 & 10584 & +97.9\% \\

Qwen-3.8-Max
& 78.2 & 1605 & -43.8\%
& 48.8 & 2507 & -65.6\%
& 30.0 & 7780 & -26.4\%
& 39.3 & 2921 & -65.0\%
& 38.4 & 1813 & -66.1\% \\

GLM-5.2
& 82.0 & 1472 & -48.5\%
& 66.9 & 2450 & -66.3\%
& 36.7 & 5524 & -47.7\%
& 43.0 & 2355 & -71.8\%
& 40.4 & 1930 & -63.9\% \\

DeepSeek-V4.1-Flash
& 84.6 & 1781 & -37.7\%
& 68.7 & 2892 & -60.3\%
& 36.0 & 5949 & -43.7\%
& 48.7 & 3371 & -59.6\%
& 38.0 & 2822 & -47.2\% \\

GPT-5.6-Luna
& 76.4 & \textbf{1326} & \textbf{-53.6}\%
& 53.0 & \textbf{1767} & \textbf{-75.7\%}
& 30.0 & \textbf{4320} & \textbf{-59.1\%}
& 37.6 & \textbf{1835} & \textbf{-78.0\%}
& 38.9 & \textbf{1665} & \textbf{-68.9\%} \\

\midrule

\rowcolor{gray!18}
\textbf{MetaCtrl}
& \textbf{88.0} & 1644 & -42.5\%
& \textbf{74.7} & 3036 & -58.3\%
& \textbf{56.7} & 5899 & -44.2\%
& \textbf{52.3} & 2965 & -64.5\%
& \textbf{42.4} & 2234 & -58.2\% \\

\bottomrule
\end{tabular}
}
\end{table*}

\begin{figure}[t]
	\centering
	\includegraphics[width=\linewidth]{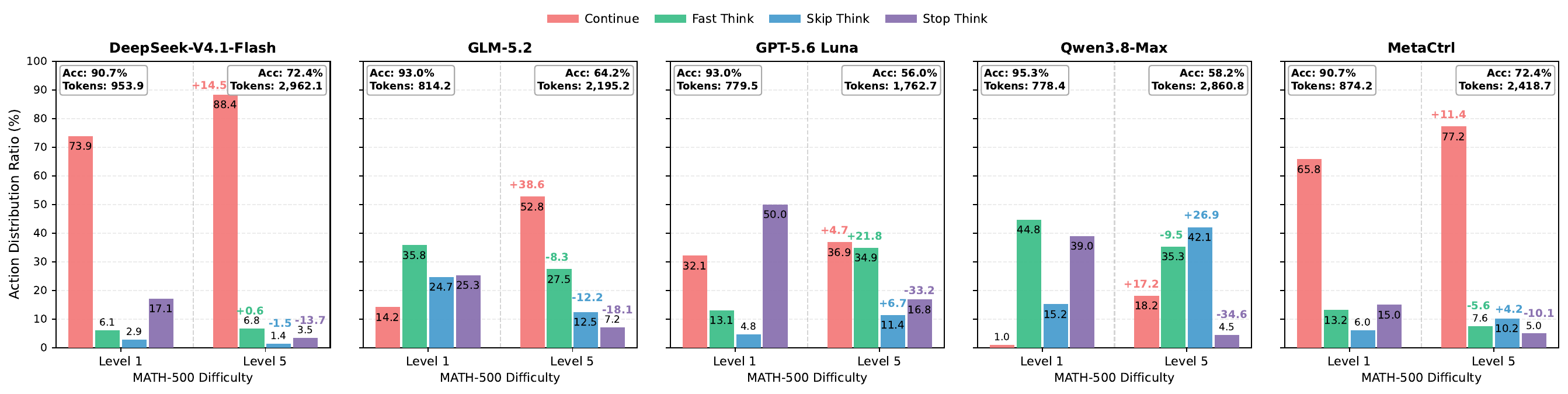}
	\caption{Action distributions of frontier LLMs as controllers and MetaCtrl across MATH-500 difficulty levels. The reasoning model is DeepSeek-R1-Distill-Qwen-7B.}
    \label{fig:api_result}
    \vspace{-0.2cm}
\end{figure}

\subsection{Main Results}

\textbf{MetaCtrl improves reasoning accuracy while reducing redundant reasoning.} As shown in Tables~\ref{tab:dsr_main_results}, on the seen DeepSeek-R1-Distill-Qwen-7B reasoner, MetaCtrl achieves the highest accuracy on all four benchmarks, improving average accuracy from 53.3\% to 59.0\% over Vanilla while reducing total generated tokens by 47.3\%. Compared with D-Prompt, TokenSkip, TH2T, and ACTS, MetaCtrl improves average accuracy by 7.0, 12.2, 4.5, and 6.7 percentage points, while achieving an additional 42.7, 12.7, 14.7, and 25.5 points of token reduction, respectively. NoThinking and CoT-Valve exhibit more aggressive compression, with 18.0 and 2.8 points greater token reduction than MetaCtrl, but at substantial accuracy costs of 11.8 and 11.0 points. Notably, ACTS is the closest external control baseline, but requires a predefined token budget that directly affects both accuracy and generation length \citep{xia2026agentic} and can be difficult to specify a priori, as the computation required for a correct solution varies across problems and reasoners. Overall, MetaCtrl achieves a consistently stronger accuracy--efficiency trade-off, suggesting that trajectory-dependent metacognitive control can reduce redundant reasoning while preserving or improving reasoning accuracy. More detailed comparisons across different baseline categories, end to end latency measurements, and case studies are provided in Appendices~\ref{app:baseline_analysis},~\ref{app:latency}, and~\ref{app:case_study}, respectively.

\begin{figure}[t]
	\centering
	\includegraphics[width=\linewidth]{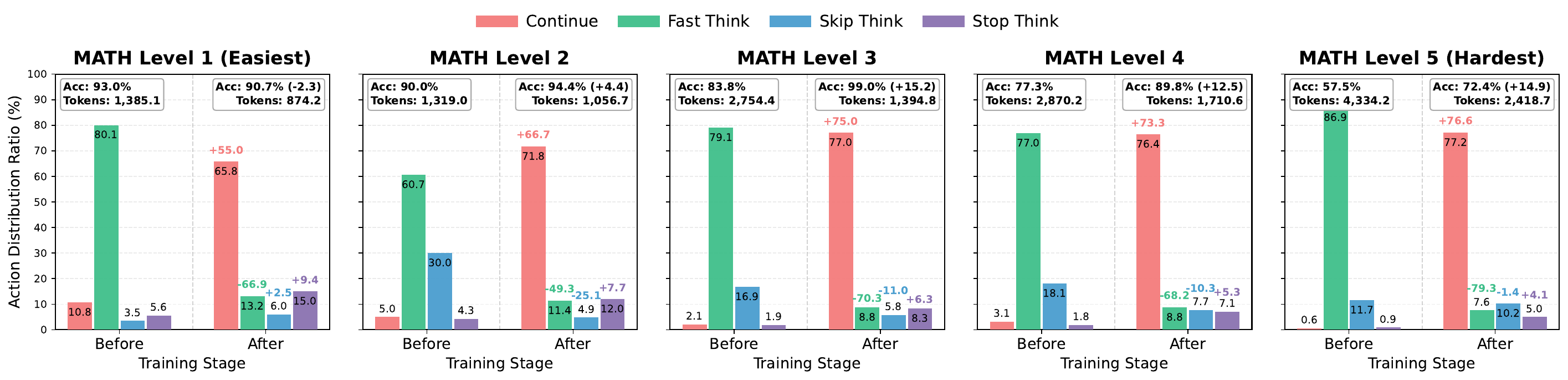}
	\caption{Action distribution change by MetaCtrl training. The reasoning model is DeepSeek-R1-Distill-Qwen-7B.}
    \label{fig:train_action}
    \vspace{-0.2cm}
\end{figure}

\textbf{MetaCtrl generalizes effectively to unseen reasoners.}
Beyond substantially reducing redundant reasoning and improving accuracy on the training-time reasoner, MetaCtrl transfers directly to unseen reasoners without any further training, while retaining strong accuracy--efficiency trade-offs. As shown in Tables~\ref{tab:qwen3_main_results}, on Qwen3-8B and Qwen3-14B, MetaCtrl improves average accuracy from 67.0\% to 70.5\% and from 71.0\% to 74.1\%, respectively, while reducing total generated tokens by 50.0\% and 53.7\%. Compared with ASAG, the strongest baseline on these reasoners, MetaCtrl achieves comparable or higher accuracy while substantially larger token reductions (50.0\% vs.\ 31.2\% on Qwen3-8B and 53.7\% vs.\ 26.9\% on Qwen3-14B). As shown in Tables~\ref{tab:dsr_main_results}, on DeepSeek-R1-Distill-Qwen-32B, MetaCtrl maintains accuracy close to Vanilla, achieving 62.1\% versus 62.3\%, while reducing generation length by 45.6\%. These results indicate that MetaCtrl learns transferable metacognitive control that generalizes across different reasoners and model scales without additional post training. Generalization to additional reasoner families is provided in Appendix~\ref{app:general_result}.

\textbf{MetaCtrl generalizes across benchmarks and domains.}
As shown in Tables~\ref{tab:dsr_main_results} and~\ref{tab:qwen3_main_results}, MetaCtrl consistently achieves strong accuracy with substantially shorter generation lengths across mathematical and scientific reasoning tasks. On MATH-500, it improves or maintains high accuracy across all four reasoners while reducing generation length by 35.2\% to 59.0\%. Similar gains are observed on more challenging mathematical benchmarks, including AIME 2024, OmniMath, and OlympiadBench, where MetaCtrl substantially shortens reasoning trajectories while maintaining competitive accuracy. The same trend extends to GPQA Diamond, where MetaCtrl achieves accuracies of 42.4\%, 53.2\%, 57.1\%, and 63.6\% across the four reasoners, with generation reductions ranging from 55.1\% to 62.5\%. These results suggest that MetaCtrl provides effective reasoning control across diverse domains and problem difficulty levels. Additional code reasoning results are reported in Appendix~\ref{app:general_result}.

\textbf{Comparison with frontier LLM controllers.}
Table~\ref{tab:api_controller_baselines} shows that stronger general purpose LLMs do not necessarily provide more effective reasoning control. MetaCtrl achieves the highest accuracy on all five benchmarks, outperforming the best frontier LLM controller by 2.0 to 20.0 percentage points. Although some frontier LLM controllers achieve larger token reductions on individual benchmarks, these gains often coincide with substantial accuracy losses. As shown in Figure~\ref{fig:api_result}, GLM-5.2, GPT-5.6-Luna, and Qwen-3.8-Max frequently apply aggressive shortening actions to difficult MATH 500 problems, coinciding with marked accuracy drops, whereas MetaCtrl better preserves continued reasoning on difficult problems while still reducing computation. These results suggest that effective control requires selectively reducing unnecessary computation while retaining the reasoning needed for correctness, rather than simply shortening trajectories.

\textbf{Adaptive Reasoning Allocation by Difficulty.} Figure~\ref{fig:api_result} examines how control policies vary with problem difficulty on MATH-500. For MetaCtrl, \textit{Continue} increases from 65.8\% at Level 1 to 77.2\% at Level 5, while \textit{Stop Think} decreases from 15.0\% to 5.0\%, suggesting that the controller becomes increasingly likely to preserve ongoing reasoning as problem difficulty increases. A similar coarse trend appears across the frontier LLM controllers, all of which increase \textit{Continue} and reduce \textit{Stop Think} as difficulty rises. Among them, DeepSeek-V4.1-Flash exhibits the most similar action distributions to MetaCtrl across difficulty levels and also achieves the strongest accuracy among the frontier LLM controllers, matching MetaCtrl at both difficulty levels but with longer generations. Beyond this shared trend, the controllers differ in their finer grained adjustments, with GPT-5.6-Luna substantially increasing \textit{Fast Think}, and Qwen3.8-Max increasing \textit{Skip Think}. Overall, this difficulty-aware control suggests that MetaCtrl intervenes more conservatively on harder problems while applying stronger reasoning compression on easier ones, contributing to its strong accuracy with shorter generations across difficulty levels. Further analyses of control behavior across tasks and reasoners are provided in Appendix~\ref{app:action_distribution}.

\textbf{MetaCtrl training learns effective reasoning control.} Figure~\ref{fig:train_action} compares the controller’s action distributions before and after training across MATH-500 difficulty levels. Before training, the controller is dominated by \textit{Fast Think}, whereas training shifts the policy strongly toward \textit{Continue} at every difficulty level. On the hardest Level 5 problems, for example, \textit{Continue} increases from 0.6\% to 77.2\%, while \textit{Fast Think} decreases from 86.9\% to 7.6\%. This policy shift improves accuracy while reducing generation length on Levels 2 through 5, with accuracy gains of 12.5 to 15.2 percentage points and generation reductions of 40.4\% to 49.4\% on the harder Levels 3 through 5. Overall, these results suggest that effective reasoning control is not achieved by simply encouraging shorter or faster reasoning, but by learning when to preserve computation and when to reduce it. Further analyses of training dynamics and policy evolution are provided in Appendix~\ref{app:training_dynamics}.

\subsection{Ablation Study}

\begin{table*}[t]
\caption{Ablation study on reward-function coefficients. The reasoning model is DeepSeek-R1-Distill-Qwen-7B.}
\label{tab:reward_ablation}
\centering
\small
\setlength{\tabcolsep}{3.8pt}
\renewcommand{\arraystretch}{1.12}

\resizebox{\textwidth}{!}{
\begin{tabular}{ccc | ccc ccc ccc ccc ccc}
\toprule

\multicolumn{3}{c|}{\textbf{Reward Setting}}
& \multicolumn{3}{c}{\textbf{MATH-500}}
& \multicolumn{3}{c}{\textbf{AMC}}
& \multicolumn{3}{c}{\textbf{AIME2024}}
& \multicolumn{3}{c}{\textbf{OlympiadBench}}
& \multicolumn{3}{c}{\textbf{GPQA Diamond}} \\
\cmidrule(lr){1-3}
\cmidrule(lr){4-6}
\cmidrule(lr){7-9}
\cmidrule(lr){10-12}
\cmidrule(lr){13-15}
\cmidrule(lr){16-18}

\textbf{$r_{\mathrm{correct}}$}
& \textbf{$r_{\mathrm{wrong}}$}
& \textbf{$\lambda_{\mathrm{eff}}$}
& \textbf{Acc.} & \textbf{Len.} & \textbf{Reduc.}
& \textbf{Acc.} & \textbf{Len.} & \textbf{Reduc.}
& \textbf{Acc.} & \textbf{Len.} & \textbf{Reduc.}
& \textbf{Acc.} & \textbf{Len.} & \textbf{Reduc.}
& \textbf{Acc.} & \textbf{Len.} & \textbf{Reduc.} \\
\midrule

\multicolumn{3}{c|}{Vanilla (No Controller)}
& 85.2 & 2857 &
& 73.5 & 7279 &
& 50.0 & 10570 &
& 48.6 & 8347 &
& 32.8 & 5349 & \\

\midrule

1.5
& -1.0
& 0.5
& 88.8 & 1739 & -39.1\%
& 76.1 & 3175 & -56.4\%
& 40.7 & 6261 & -40.8\%
& 52.3 & 3092 & -63.0\%
& \textbf{43.4} & 2789 & -47.9\% \\

2.0
& -1.0
& 0.5
& \textbf{90.8} & 1921 & -32.8\%
& \textbf{77.6} & 4162 & -42.8\%
& 44.0 & 8106 & -23.3\%
& \textbf{53.3} & 3870 & -53.6\%
& 37.9 & 4818 & -9.9\% \\

1.0
& -0.5
& 0.5
& 85.8 & 1553 & -45.6\%
& 71.1 & 2545 & -65.0\%
& 40.7 & 5013 & -52.6\%
& 45.9 & 2590 & -69.0\%
& 36.9 & 1998 & -62.6\% \\

1.0
& -1.0
& 1.0
& 82.0 & \textbf{1284} & -55.1\%
& 59.0 & \textbf{2053} & -71.8\%
& 30.7 & \textbf{4626} & -56.2\%
& 40.9 & \textbf{1968} & -76.4\%
& 37.5 & \textbf{1463} & -72.6\% \\

\midrule

% Default setting
\rowcolor{gray!18}
\textbf{1.0}
& \textbf{-1.0}
& \textbf{0.5}
& 88.0 & 1644 & -42.5\%
& 74.7 & 3036 & -58.3\%
& \textbf{56.7} & 5899 & -44.2\%
& 52.3 & 2965 & -64.5\%
& 42.4 & 2234 & -58.2\% \\

\bottomrule
\end{tabular}
}
\vspace{-0.25cm}
\end{table*}

\textbf{Reward coefficients.}
Table~\ref{tab:reward_ablation} examines the effect of reward coefficients on accuracy and generation length. The default setting, $r_{\mathrm{correct}}=1.0$, $r_{\mathrm{wrong}}=-1.0$, and $\lambda_{\mathrm{eff}}=0.5$, is the only tested configuration that improves accuracy over Vanilla on all five benchmarks, while reducing generation length by 42.5\% to 64.5\%. Weakening the wrong answer penalty to $-0.5$ yields stronger compression but lowers accuracy on several tasks, while increasing $\lambda_{\mathrm{eff}}$ to 1.0 further shortens generation at a larger accuracy cost. Increasing $r_{\mathrm{correct}}$ benefits some tasks but is less consistent across benchmarks. Overall, these results highlight the importance of balancing correctness preservation and reasoning compression. Additional ablation results are provided in Appendix~\ref{app:ablation_studies}.

\section{Conclusion}
In this paper, we present MetaCtrl, a lightweight metacognitive controller that dynamically regulates a frozen reasoner based on its evolving reasoning trajectory. MetaCtrl learns when and how to intervene through GRPO with a correctness-gated efficiency objective, without supervised intervention trajectories, predefined reasoning budgets, or reasoner retraining. Extensive experiments across seven benchmarks show that MetaCtrl consistently improves the reasoning accuracy of controlled LRMs while substantially reducing their reasoning length, and generalizes effectively to unseen LRMs spanning different model scales and architectures as well as across diverse reasoning domains. With matched GPU memory, MetaCtrl also substantially reduces end to end LRM inference latency. These results demonstrate that decoupling metacognitive control from problem solving provides an effective and transferable mechanism for adaptively allocating reasoning computation.

\subsection*{AI Use Statement}
Generative AI tools were used solely to assist with language editing and writing refinement, including improving grammar, clarity, readability, and presentation of the manuscript. They were not used to develop the research methodology, design experiments, generate or analyze experimental results, or formulate the scientific conclusions of this work. All AI-assisted text was carefully reviewed, revised, and verified by the authors to ensure that it accurately reflects the intended technical content. The authors take full responsibility for the final content of the paper, including all statements, claims, and conclusions.

\bibliography{iclr2027_conference}
\bibliographystyle{iclr2027_conference}

\newpage
\appendix

\startcontents[appendices]
\section*{Table of Contents}
\printcontents[appendices]{}{1}{
    \setcounter{tocdepth}{2}
}
\clearpage

\section{Related Work Details}
\label{app:related_work}

We provide a more detailed discussion of the efficient-reasoning paradigms summarized in Figure~\ref{fig:related_work}.

\paragraph{Prompt-based methods.}
Prompt-based approaches improve reasoning efficiency without modifying model parameters \citep{lee2025well,yu2025premise,zhang2025long}. Concise Chain-of-Thought \citep{renze2024benefits} and Chain of Draft \citep{xu2025chain} prompt reasoning models to produce shorter and more concise reasoning traces, while TALE \citep{han2025token} estimates question-specific token budgets before generation. Sketch-of-Thought \citep{aytes2025sketch}, ThinkSwitcher \citep{liang2025thinkswitcher}, and DiffAdapt \citep{liu2026diffadapt} further adapt reasoning formats or modes according to input characteristics. These methods can adapt computation to the input, but their main control decisions are made before or at the beginning of generation, when it is difficult to determine from the question alone how much computation is needed or how reasoning should proceed. Moreover, such decisions cannot respond to later changes in reasoning progress as the solution develops. MetaCtrl instead conditions its interventions on the evolving reasoning trace and adapts computation online throughout generation.

\paragraph{Model-based methods.}
Model-based approaches internalize efficient reasoning through post-training \citep{kang2025c3ot,ma2025cot,yang2025think}. System-2 distillation \citep{yu2024distilling} and TokenSkip \citep{xia2025tokenskip} compress deliberative reasoning, while O1-Pruner \citep{luo2026o1}, ThinkPrune \citep{hou2025thinkprune}, and L1 \citep{aggarwal2025l1} optimize reasoning length through reinforcement learning or explicit token constraints. DAST \citep{shen2025dast} and TH2T \citep{liu2026think} further incorporate difficulty or redundancy awareness to adapt computation across problems. These methods modify the reasoner itself and therefore require model-specific post-training, coupling problem solving with computation control. This is particularly costly for difficulty-aware methods because perceived problem difficulty is relative to the capability of the underlying reasoner, potentially requiring adaptation when the reasoner changes. MetaCtrl instead freezes the reasoner and trains only a lightweight meta-level controller, decoupling reasoning control from problem solving and avoiding costly reasoner training. This advantage becomes more pronounced as reasoner scale grows, while also enabling transfer across different reasoners without retraining them.

\paragraph{Output-based methods.}
Output-based approaches regulate the ongoing generation process or intervene on its internal representations \citep{liao2025reward,liu2025answer,tikhonov2026confidence}. Coconut \citep{hao2024training} performs reasoning in continuous latent states rather than discrete language tokens. DEER \citep{yang2026dynamic} and Certaindex \citep{fu2025efficiently} use confidence or answer-stability signals to determine when additional computation is unnecessary, while SEAL \citep{chen2025seal} intervenes through internal representations. ASAG \citep{li2026stop} combines model confidence and attention entropy to adapt generation strategies. These approaches often rely on predefined decision rules or thresholds, or require access to internal states such as hidden representations, attention, or logits. Moreover, fixed confidence-based criteria can behave differently across reasoning states and problem difficulty. MetaCtrl instead learns when and how to intervene directly from correctness--efficiency rewards using only the observable reasoning trajectory, without hand-crafted control criteria or access to internal model states.

\paragraph{External reasoning control.}
External control methods regulate a reasoner using a separate model. SpecReason \citep{pan2025specreason} and Speculative Thinking \citep{yang2025speculative} use stronger models to verify, correct, or guide weaker reasoners. Although this can reduce computation performed by the target reasoner, the external model participates directly in problem solving by providing solution-specific reasoning guidance, effectively coupling inference across multiple capable agents and increasing system complexity. CGI \citep{wang2026intervene} injects external state-based feedback from humans or LLM proxies. Its evaluator assesses the rationality and completeness of the current reasoning and can provide task-specific corrective guidance, thereby participating in the problem solving process to some extent. When an LLM proxy is used, this effectively forms a multi-model problem-solving system in which a stronger external model guides the target reasoner. CGI additionally trains the reasoner to respond to such interventions and may rely on an LLM proxy substantially larger than the target reasoner. ACTS \citep{xia2026agentic} is more closely related to our setting because it controls a frozen reasoner with a separate controller. However, ACTS relies on distilled SFT data and requires a predefined token budget, which can be difficult to determine a priori because the required computation depends jointly on problem difficulty and reasoner capability.

MetaCtrl differs from these approaches in three respects. First, it keeps the target reasoner fully frozen and requires no model-specific adaptation. Second, its controller does not participate in solving the task or provide solution-specific content. Instead, it selects a small set of task-agnostic control actions that determine how the reasoner should proceed. The controller is learned directly through reinforcement learning from evolving reasoning trajectories, without supervised intervention trajectories or distilled SFT data. Third, MetaCtrl requires neither a predefined reasoning budget nor a large external proxy, allowing a lightweight controller to dynamically regulate when and how reasoning proceeds as the solution unfolds.

\section{Prompt Templates}
\label{app:prompt_template}

\subsection{Intervention Prompts}
\label{app:intervention_prompts}

MetaCtrl realizes its meta-level actions through lightweight textual prompts inserted into the reasoner's context at each control point. Table~\ref{tab:intervention_prompts} reports the exact prompts used in our experiments. \textsc{Continue} introduces no additional text, leaving the reasoning context unchanged. For the other actions, the corresponding prompt is appended to the current context before the frozen reasoner resumes generation. The leading newline used in implementation is omitted from the table for readability.

\begin{table}[h]
\caption{Textual intervention prompts associated with the MetaCtrl action space.}
\centering
\small
\begin{tabular}{p{0.16\linewidth} p{0.76\linewidth}}
\toprule
\textbf{Action} & \textbf{Intervention Prompt} \\
\midrule

\textsc{Continue}
&
No textual intervention. \\[2mm]

\textsc{Fast Think}
&
\texttt{Let's keep only the essential derivation, avoid repetition, and proceed efficiently.} \\[2mm]

\textsc{Skip Think}
&
\texttt{Let's solve this directly with the minimum necessary reasoning and move to the answer.} \\[2mm]

\textsc{Stop Think}
&
\texttt{The reasoning is sufficient; state only the final conclusion briefly and close the reasoning now.} \\

\bottomrule
\end{tabular}
\label{tab:intervention_prompts}
\end{table}

Importantly, \textsc{Stop Think} does not forcibly truncate generation. Instead, it instructs the frozen reasoner to conclude its current reasoning and produce the final answer, preserving the same text-based interaction interface used by the other interventions.

\subsection{Controller Prompt}
The controller receives the original problem and the reasoner's latest reasoning trajectory. Its system prompt is:

\begin{quote}
\small
You control a frozen mathematical reasoner while it solves one problem.

The reasoner always generates its first reasoning step without controller
intervention. You are called only if reasoning is still active after that
step. On every controller turn, you receive the reasoner's latest step and
select one action from \texttt{continue}, \texttt{fast\_think},
\texttt{skip\_think}, or \texttt{stop\_think}. Optimize for a correct
answer first; among correct solutions, prefer the shortest sufficient
reasoning.

\textbf{Actions:}

\begin{itemize}
    \item \texttt{continue}: add no intervention and let the reasoner
    continue naturally.
    \item \texttt{fast\_think}: ask for only the essential derivation and
    no repetition.
    \item \texttt{skip\_think}: skip redundant intermediate work and move
    to the shortest remaining derivation.
    \item \texttt{stop\_think}: run one bounded conclusion continuation,
    close the thinking channel, and proceed to the final answer. Existing
    reasoning is preserved and never truncated.
\end{itemize}

Output exactly one line and no explanation:

\texttt{Action:
<continue|fast\_think|skip\_think|stop\_think>}
\end{quote}

\section{Experimental Details}
\label{app:experimental_details}

\subsection{Datasets and Metrics.} We use 7,500 mathematical problems sampled from OpenR1-Math \citep{openr1,xia2026agentic} for online controller training. We evaluate across three reasoning domains: mathematics on MATH-500 \citep{hendrycks2021measuring}, AIME 2024 \citep{maaAIME}, Omni-MATH \citep{gao2025omni}, OlympiadBench \citep{he2024olympiadbench}, and AMC \citep{aimo2024amc}, science on GPQA Diamond \citep{rein2024gpqa}, and code on LiveCodeBench \citep{jain2025livecodebench}. We report accuracy (pass@1), average token length, and length reduction ratio, which denotes the percentage reduction in token length relative to Vanilla. For controlled methods, token length includes generations from both the target reasoner and the controller.

\subsection{Backbone Models.} During training, we initialize the controller from Qwen3-4B-Instruct-2507 \citep{yang2025qwen3} and pair it with a frozen DeepSeek-R1-Distill-Qwen-7B reasoner \citep{guo2025deepseek}. At evaluation, we test the controller on the training-time reasoner and assess cross-reasoner generalization by directly transferring it, without further training, to three unseen reasoners: DeepSeek-R1-Distill-Qwen-32B, DeepSeek-R1-Distill-Llama-8B \citep{guo2025deepseek}, Qwen3-8B, and Qwen3-14B \citep{yang2025qwen3}.

\subsection{Baseline Details}
\label{app:baseline_details}

We compare against a broad set of recent and strong baselines across six categories:
(i) \textbf{Vanilla}: standard reasoning without intervention;
(ii) \textbf{Prompt-based methods}: D-Prompt \citep{liu2026think},
NoThinking \citep{ma2025reasoning}, and TALE \citep{han2025token};
(iii) \textbf{Model-based methods}: TokenSkip \citep{xia2025tokenskip},
CoT-Valve \citep{ma2025cot}, AdaCtrl \citep{huang2025adactrl} and TH2T \citep{liu2026think};
(iv) \textbf{Output-based methods}: Dynasor \citep{fu2025efficiently},
DEER \citep{yang2026dynamic}, and ASAG \citep{li2026stop}; and
(v) \textbf{External reasoning control}: ACTS \citep{xia2026agentic}; and
(vi) \textbf{Frontier LLM as controller}: GLM-5.2 \citep{zeng2026glm}, Qwen3.8-Flash, Qwen3.8-Max \citep{qwen38}, DeepSeek-V4.1-Flash \citep{xu2026deepseek}, and GPT-5.6-Luna\footnote{\url{https://developers.openai.com/api/docs/models/gpt-5.6-luna}},, each accessed through its official API and directly used as the controller under the same controller--reasoner interface.

\paragraph{Prompt-based methods.} \textbf{D-Prompt} \citep{liu2026think} augments the input with a difficulty reminder and encourages the reasoner to adjust its response length according to its self-assessed problem difficulty. \textbf{NoThinking} \citep{ma2025reasoning} instructs the model to bypass explicit intermediate reasoning and directly generate the answer. \textbf{TALE} \citep{han2025token} controls reasoning length through token-budget prompting, assigning a constrained reasoning budget before generation.

\paragraph{Model-based methods.} \textbf{TokenSkip} \citep{xia2025tokenskip} fine-tunes the target reasoner to skip non-essential reasoning tokens during generation. \textbf{CoT-Valve} \citep{ma2025cot} learns controllable directions in model parameter space to regulate chain-of-thought length. \textbf{TH2T} \citep{liu2026think} applies two-stage fine-tuning to induce difficulty and redundancy awareness, enabling adaptive reasoning depth while suppressing redundant reasoning. \textbf{AdaCtrl} \citep{huang2025adactrl} combines difficulty-aware fine-tuning with reinforcement learning to self-assess problem difficulty and adaptively allocate reasoning budgets.

\paragraph{Output-based methods.}
\textbf{Dynasor} \citep{fu2025efficiently} checks partial reasoning at regular intervals and stops generation once successive intermediate answers reach sufficient agreement. \textbf{DEER} \citep{yang2026dynamic} instead evaluates intermediate predictions at adaptive transition points in the reasoning process and terminates early when a high-confidence answer emerges. \textbf{ASAG} \citep{li2026stop} jointly uses model confidence and attention-state information to determine whether further reasoning is likely to be useful and adaptively controls generation.

\paragraph{External reasoning control.} \textbf{ACTS} \citep{xia2026agentic} employs a separately trained controller to steer a frozen reasoner step by step. Its controller conditions on the evolving reasoning trajectory and an explicit remaining token budget, and is first initialized through supervised fine-tuning on steering data constructed from expert reasoning trajectories, followed by budget-conditioned reinforcement learning.

For the baseline methods, we directly report the results from the original TH2T \citep{liu2026think} and ASAG \citep{li2026stop} papers. For all evaluations of our method, we adopt the same configurations used in these prior works, ensuring fair comparison under matched inference settings. For ACTS \citep{xia2026agentic}, we reproduce its evaluation using greedy decoding for the reasoner to ensure a consistent decoding protocol across baselines. Implementation details are described in the Appendix~\ref{app:implementation}.

\subsection{Implementation Details}
\label{app:implementation}

\paragraph{Training and Serving.} We directly train the controller with GRPO, without supervised fine-tuning. We use a learning rate of $1\times10^{-6}$, a GRPO group size of 8, a rollout batch size of 32, and a global batch size of 64. We set $r_{\mathrm{correct}}=1$, $r_{\mathrm{wrong}}=-1$, and the efficiency coefficient $\lambda_{\mathrm{eff}}=0.5$. During training, the controller is sampled with temperature 1.0 and top-$p$ 0.95, whereas the frozen reasoner uses greedy decoding. All experiments run on 8 NVIDIA H20 GPUs. RL training is implemented with SLIME \citep{slime_github}. At inference time, the controller and reasoner are hosted as separate SGLang \citep{zheng2024sglang} servers.

\paragraph{Evaluation Protocol.} For fair comparison, we directly report the baseline results from the original TH2T \citep{liu2026think} and ASAG \citep{li2026stop} papers, and evaluate our method using the same configurations adopted in these works across all experiments. Since TH2T and ASAG report results with greedy reasoner decoding, we also reproduce ACTS under greedy reasoner decoding rather than directly using its official evaluation setting, which samples from the reasoner with temperature $0.6$ and top-$p$ $0.95$. Since ACTS requires an explicit input reasoning budget, we set the token budget to 2,000 for MATH-500, 5,000 for AIME 2024, and 4,000 for Omni-MATH, OlympiadBench, and GPQA Diamond. For our method, the controller follows the ACTS decoding configuration with temperature $0.7$ and top-$p$ $0.8$, while the reasoner uniformly uses greedy decoding across all evaluations. The reasoner-side generation limits are matched to those of the corresponding baseline settings to avoid confounding performance differences with inference configuration.

\section{Detailed Comparison with Efficient Reasoning Baselines}
\label{app:baseline_analysis}

Our main results are presented in Table~\ref{tab:dsr_main_results} and Table~\ref{tab:qwen3_main_results}. We provide a more detailed comparison with the four major families of efficient reasoning methods considered in our experiments. We focus on both empirical performance and how each paradigm allocates test-time computation.

\paragraph{Prompt-based methods.}
Prompt-based approaches regulate reasoning through instructions or token budgets specified before generation, and therefore cannot adapt their decisions to how the reasoning trajectory unfolds. This limitation is reflected in our results. On DeepSeek-R1-Distill-Qwen-7B, D-Prompt achieves 52.0\% average accuracy with only a 4.6\% token reduction, whereas MetaCtrl reaches 59.0\% accuracy while reducing generation by 47.3\%. NoThinking obtains a larger reduction of 65.3\%, but its average accuracy drops to 47.2\%, illustrating the cost of uniformly suppressing deliberation. The same trend holds for other reasoning models: compared with TALE, MetaCtrl improves average accuracy from 65.0\% to 70.5\% on Qwen3-8B and from 70.2\% to 74.1\% on Qwen3-14B, while increasing token reduction from 26.0\% to 50.0\% and from 17.5\% to 53.7\%, respectively. These results suggest that the amount of useful computation is difficult to determine from the input alone and is better adjusted according to the evolving reasoning state.

\paragraph{Model-based methods.}
Model-based methods improve reasoning efficiency by modifying the target reasoner itself through supervised fine-tuning or reinforcement learning. Such approaches can learn adaptive reasoning behaviors, but couple computation control with problem solving and generally require model-specific post-training. In contrast, MetaCtrl keeps the reasoner frozen and learns a separate metacognitive policy. On the DeepSeek-R1-Distill-Qwen-7B reasoner, TH2T, the strongest model-based baseline in average accuracy, achieves 54.5\% accuracy with a 32.6\% token reduction, while MetaCtrl reaches 59.0\% accuracy and a 47.3\% reduction. TokenSkip and CoT-Valve achieve 46.8\% and 48.0\% average accuracy, respectively, despite reducing generation by 34.6\% and 50.1\%.

A further limitation of difficulty-aware model-based approaches such as TH2T is that perceived problem difficulty is inherently reasoner-dependent. Models at different scales can have substantially different capability boundaries, such that the same problem may be easy for a stronger reasoner but difficult for a weaker one. Consequently, a difficulty criterion calibrated for one model or benchmark does not necessarily transfer faithfully to another. Applying a shared difficulty standard across reasoners therefore risks conflating task difficulty with model capability, whereas properly calibrating difficulty would require model- and benchmark-specific estimation. Moreover, TH2T requires constructing additional training data and post-training each target reasoner, further increasing the adaptation cost.

MetaCtrl avoids these requirements by estimating how much further computation is useful directly from the evolving reasoning trajectory of the current reasoner. More importantly, the controller is trained once as a smaller meta-level model and can then regulate multiple larger frozen reasoners without retraining them individually. This substantially reduces the cost of deploying efficient reasoning across a family of models: rather than separately collecting data and post-training every target reasoner, a single reusable controller can improve the inference efficiency of multiple larger models. Empirically, on DeepSeek-R1-Distill-Qwen-32B, the same controller achieves 62.1\% average accuracy with a 45.6\% token reduction, compared with 61.3\% and 36.4\% for TH2T. This decoupling enables a single controller to generalize across reasoners, eliminating the need for model-specific retraining of computation control.

\paragraph{Output-based methods.}
Output-based methods regulate computation during inference and are therefore more adaptive than static prompting. However, existing approaches typically rely on predefined stopping criteria, confidence estimates, agreement statistics, or privileged model-internal signals. MetaCtrl instead learns its intervention policy directly from trajectory-level correctness and efficiency outcomes and requires only the observable reasoning trace. On Qwen3-8B, ASAG, the strongest output-based baseline, obtains 69.9\% average accuracy with a 31.2\% token reduction, whereas MetaCtrl achieves 70.5\% accuracy while reducing generation by 50.0\%. On Qwen3-14B, MetaCtrl achieves comparable accuracy to ASAG (74.1\% vs.\ 74.0\%) but more than doubles the token reduction, from 26.9\% to 53.7\%. Compared with DEER and Dynasor, MetaCtrl likewise achieves higher average accuracy together with substantially larger reductions in generated tokens. These results indicate that learning trajectory-dependent interventions can provide more aggressive computation savings without relying on manually specified control signals. More broadly, the fixed criteria underlying many output-based methods may not transfer uniformly across reasoners with different capability boundaries or across problems with different computational demands. MetaCtrl instead conditions each control decision on the evolving reasoning trajectory, allowing the regulation strategy to adapt dynamically to both the current reasoner and its problem solving progress.

\paragraph{External reasoning control.}
ACTS is the most closely related baseline because it also separates the controller from a frozen reasoner and makes online decisions from the evolving trajectory. However, ACTS requires supervised steering data and explicitly conditions the controller on a predefined remaining token budget. MetaCtrl removes both requirements: it is trained directly from outcome-level reinforcement and determines how much additional reasoning is useful without a user-specified budget. The performance gap is consistent across all four evaluated reasoners. On DeepSeek-R1-Distill-Qwen-7B, MetaCtrl improves average accuracy from 52.3\% to 59.0\% and increases token reduction from 21.8\% to 47.3\%. On the unseen 32B reasoner, it improves accuracy from 57.8\% to 62.1\% while increasing token reduction from 11.7\% to 45.6\%. Similar gains are observed on Qwen3-8B and Qwen3-14B, where MetaCtrl improves average accuracy over ACTS by 7.4 and 7.6 percentage points, respectively, while also achieving larger token reductions (50.0\% vs.\ 44.1\% and 53.7\% vs.\ 39.5\%). Although increasing the prescribed budget in ACTS may improve accuracy, the minimum budget sufficient for a correct solution is unknown a priori and varies across problems and reasoners, leaving users to choose it heuristically. MetaCtrl instead learns to allocate additional computation online from the evolving reasoning trajectory without requiring such a predefined budget.

Across the four paradigms, the results reveal complementary limitations of existing approaches: prompt-based methods decide computation too early, model-based methods require modifying each target reasoner, output-based methods rely on predefined control signals, and existing external controllers require additional supervision or explicit reasoning budgets. MetaCtrl instead learns a separate, budget-free, trajectory-conditioned control policy directly from correctness and efficiency outcomes while keeping the reasoner frozen. Rather than requiring a predefined computation budget, it dynamically determines how much additional reasoning is needed based on the evolving trajectory. This design yields a better accuracy--efficiency trade-off on the training-time reasoner and, critically, transfers to unseen reasoners without additional post-training.

\section{Additional Experimental Results and Analysis}
\label{app:more_result}

\begin{table*}[t]
\caption{Performance and reasoning efficiency across unseen reasoners on AMC and LiveCodeBench.}
\label{tab:more_result}
\centering
\small
\setlength{\tabcolsep}{6.0pt}
\renewcommand{\arraystretch}{1.12}

\resizebox{0.6\textwidth}{!}{
\begin{tabular}{l ccc ccc}
\toprule

\multirow{2}{*}{\textbf{Methods}}
& \multicolumn{3}{c}{\textbf{AMC}}
& \multicolumn{3}{c}{\textbf{LiveCodeBench}} \\
\cmidrule(lr){2-4}
\cmidrule(lr){5-7}

& \textbf{Acc.$\uparrow$}
& \textbf{Len.$\downarrow$}
& \textbf{Reduc.$\uparrow$}
& \textbf{Acc.$\uparrow$}
& \textbf{Len.$\downarrow$}
& \textbf{Reduc.$\uparrow$} \\
\midrule

% ============== DeepSeek-R1-Distill-Qwen-7B ==============
\multicolumn{7}{c}{
    \textit{\textbf{DeepSeek-R1-Distill-Qwen-7B [Seen Reasoner]}}
} \\
\midrule

Vanilla
& 73.5 & 7279 &
& 38.4 & 10,516 & \\

\rowcolor{gray!18}
MetaCtrl (ours)
& 74.7 & 3036 & -58.3\%
& 42.3 & 2,932 & -72.1\% \\

\midrule

% ==================== Qwen3-8B ====================
\multicolumn{7}{c}{
    \textit{\textbf{Qwen3-8B [Unseen Reasoner]}}
} \\
\midrule

Vanilla
& 78.8 & 9083 &
& 64.7 & 8,923 & \\

\rowcolor{gray!18}
MetaCtrl (ours)
& 81.9 & 4623 & -49.1\%
& 65.0 & 5855 & -34.4\% \\

\midrule

% ==================== Qwen3-14B ====================
\multicolumn{7}{c}{
    \textit{\textbf{Qwen3-14B [Unseen Reasoner]}}
} \\
\midrule

Vanilla
& 81.9 & 8505 &
& 72.6 & 8,112 & \\

\rowcolor{gray!18}
MetaCtrl (ours)
& 84.3 & 3938 & -53.7\%
& 74.7 & 5416 & -33.2\% \\

\midrule

% ==================== DeepSeek-R1-Distill-Llama-8B ====================
\multicolumn{7}{c}{
    \textit{\textbf{DeepSeek-R1-Distill-Llama-8B [Unseen Reasoner]}}
} \\
\midrule

Vanilla
& 67.5 & 7982 &
& 41.5 & 11029 & \\

\rowcolor{gray!18}
MetaCtrl (ours)
& 68.7 & 3357 & -57.9\%
& 43.1 & 3861 & -65.0\% \\

\bottomrule
\end{tabular}
}
\end{table*}

\begin{table*}[t]
\caption{
Results of DeepSeek-R1-Distill-Llama-8B across five reasoning benchmarks.
\textbf{Acc.} denotes pass@1 accuracy, \textbf{Len.} denotes the average total generation length,
and \textbf{Reduc.} denotes the relative reduction in generation length compared with vanilla reasoning.
}
\label{tab:llama8b_results}
\centering
\small
\setlength{\tabcolsep}{3.5pt}
\renewcommand{\arraystretch}{1.12}

\resizebox{\textwidth}{!}{
\begin{tabular}{l ccc ccc ccc ccc}
\toprule

\multirow{2}{*}{\textbf{Methods}}
& \multicolumn{3}{c}{\textbf{MATH-500}}
& \multicolumn{3}{c}{\textbf{AIME 2024}}
& \multicolumn{3}{c}{\textbf{OlympiadBench}}
& \multicolumn{3}{c}{\textbf{GPQA Diamond}} \\

\cmidrule(lr){2-4}
\cmidrule(lr){5-7}
\cmidrule(lr){8-10}
\cmidrule(lr){11-13}

& \textbf{Acc.$\uparrow$}
& \textbf{Len.$\downarrow$}
& \textbf{Reduc.$\uparrow$}

& \textbf{Acc.$\uparrow$}
& \textbf{Len.$\downarrow$}
& \textbf{Reduc.$\uparrow$}

& \textbf{Acc.$\uparrow$}
& \textbf{Len.$\downarrow$}
& \textbf{Reduc.$\uparrow$}

& \textbf{Acc.$\uparrow$}
& \textbf{Len.$\downarrow$}
& \textbf{Reduc.$\uparrow$} \\

\midrule

\multicolumn{13}{c}{
    \textit{\textbf{DeepSeek-R1-Distill-Llama-8B [Unseen Reasoner]}}
} \\

\midrule

Vanilla
& 87.6 & 3928 & 
& 40.0 & 13472 & 
& 47.3 & 8416 & 
& 26.3 & 9655 &  \\

\rowcolor{gray!18}
MetaCtrl (ours)
& 89.0 & 1942 & -50.6\%
& 40.0 & 7642 & -43.3\%
& 50.1 & 3819 & -54.6\%
& 37.4 & 3192 & -66.9\% \\

\bottomrule
\end{tabular}
}
\end{table*}

\subsection{Generalization Across Benchmarks, Domains, and Reasoning Model Architectures.}
\label{app:general_result}

We further evaluate MetaCtrl on additional benchmarks, the code reasoning domain, and an unseen reasoner architecture. As shown in Tables~\ref{tab:more_result} and~\ref{tab:llama8b_results}, the same controller transfers to unseen reasoners on AMC and LiveCodeBench without additional training, reducing generation length by \(33.2\%\) to \(65.0\%\) while maintaining or improving accuracy in all settings. This transfer extends to DeepSeek-R1-Distill-Llama-8B, whose Llama backbone differs from the Qwen-based reasoner used for controller training. On AMC and LiveCodeBench, MetaCtrl improves accuracy from \(67.5\%\) to \(68.7\%\) and from \(41.5\%\) to \(43.1\%\), while reducing generation length by \(57.9\%\) and \(65.0\%\), respectively. Across MATH-500, AIME 2024, OlympiadBench, and GPQA Diamond, it further reduces generation length by \(43.3\%\) to \(66.9\%\) while matching or improving Vanilla accuracy, including an \(11.1\) percentage point gain on GPQA Diamond. Together, these results show that MetaCtrl is not tied to the training benchmark, reasoning domain, or reasoner architecture, but transfers effectively across unseen tasks and reasoning models without additional training.

\subsection{Adaptive Control Emerges from Task--Reasoner Interactions.}
\label{app:action_distribution}

\begin{figure}[t]
	\centering
	\includegraphics[width=\linewidth]{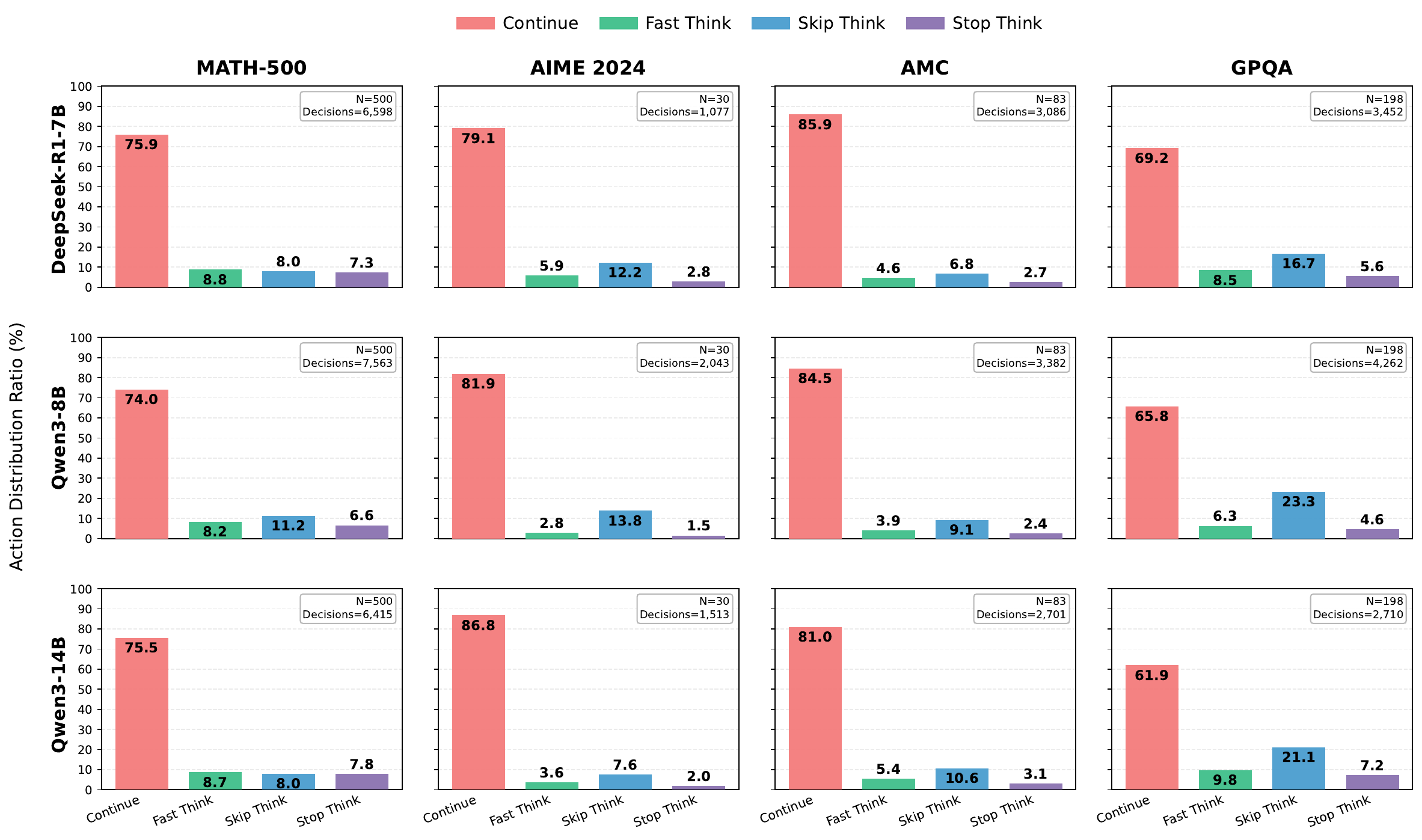}
	\caption{Controller action distributions across different reasoners and benchmarks. Each row corresponds to a reasoner and each column to a benchmark. The learned policy exhibits distinct task--reasoner-specific control patterns rather than a uniform intervention strategy. DeepSeek-R1-7B is used as shorthand for DeepSeek-R1-Distilled-Qwen2.5-7B.}
    \label{fig:action_distribution_reasoner_by_benchmark}
\end{figure}

\paragraph{Adaptive control across tasks and reasoners.}
Figure~\ref{fig:action_distribution_reasoner_by_benchmark} shows that MetaCtrl maintains a selective control policy across 12 reasoner and benchmark combinations. \textit{Continue} remains the dominant action, accounting for 61.9\% to 86.8\% of decisions, while \textit{Stop Think} is used only 1.5\% to 7.8\% of the time. \textit{Skip Think} is more frequent than \textit{Stop Think} in every setting and is the most common intervention action in 10 of the 12 combinations. These statistics indicate that the controller more frequently applies local reasoning compression than explicit termination at individual control points. At the same time, intervention rates vary systematically across tasks. GPQA elicits substantially more intervention than AIME 2024 and AMC across all three reasoners, while MATH 500, despite being relatively easier, consistently shows lower \textit{Continue} rates than AIME 2024 and AMC. This suggests that the controller is not governed by benchmark difficulty alone, but adapts to task specific reasoning patterns, intervening more when shorter or locally compressed reasoning may suffice and preserving longer reasoning when sustained computation remains useful. Overall, these patterns indicate that MetaCtrl largely preserves the reasoner’s native computation while adapting both the degree and form of intervention across tasks and reasoners.

\paragraph{Control demand is not explained by task difficulty or trajectory length alone.}
GPQA consistently induces the highest intervention rates across all three reasoners, reaching 30.8\%, 34.2\%, and 38.1\%, compared with 20.9\%, 18.1\%, and 13.2\% on AIME 2024. Importantly, this difference cannot be attributed simply to longer reasoning trajectories. Although all three reasoners receive substantially more controller decisions per problem on AIME 2024 than on GPQA, the controller intervenes considerably less often on AIME. These results suggest that control demand is not determined by task difficulty or trajectory length alone. Instead, the learned policy appears sensitive to the evolving reasoning trajectory, preserving continued computation in some settings while intervening more frequently in others.

\paragraph{Control demand depends on the model-task pair, not model scale alone.}
Increasing the reasoner size from Qwen3-8B to Qwen3-14B does not produce a consistent change in intervention rate. It decreases from 18.1\% to 13.2\% on AIME 2024, but increases from 34.2\% to 38.1\% on GPQA and from 15.5\% to 19.0\% on AMC. Meanwhile, MATH-500 remains nearly unchanged across DeepSeek-R1-7B, Qwen3-8B, and Qwen3-14B, with intervention rates of 24.1\%, 26.0\%, and 24.5\%, respectively. These non-monotonic trends suggest that control demand depends on the interaction between the reasoner and the task rather than on model scale alone. This motivates adaptive external control, since a fixed token budget or uniform stopping rule cannot directly account for such variation across reasoner-task combinations.

Overall, these results provide behavioral evidence for the central motivation of adaptive, budget-free reasoning control: useful computation should be regulated according to the evolving reasoning state, rather than predetermined from model size, task difficulty, or elapsed computation.

\begin{table*}[t]
\caption{Ablation studies on intervention position, action space, and controller initialization. The reasoning model is DeepSeek-R1-Distill-Qwen-7B.}
\label{tab:controller_ablation}
\centering
\small
\setlength{\tabcolsep}{3.8pt}
\renewcommand{\arraystretch}{1.12}

\resizebox{\textwidth}{!}{
\begin{tabular}{ll | ccc ccc ccc ccc ccc}
\toprule

\multicolumn{2}{c|}{\textbf{Ablation Setting}}
& \multicolumn{3}{c}{\textbf{MATH-500}}
& \multicolumn{3}{c}{\textbf{AMC}}
& \multicolumn{3}{c}{\textbf{AIME2024}}
& \multicolumn{3}{c}{\textbf{OlympiadBench}}
& \multicolumn{3}{c}{\textbf{GPQA Diamond}} \\
\cmidrule(lr){1-2}
\cmidrule(lr){3-5}
\cmidrule(lr){6-8}
\cmidrule(lr){9-11}
\cmidrule(lr){12-14}
\cmidrule(lr){15-17}

\textbf{Component}
& \textbf{Setting}
& \textbf{Acc.} & \textbf{Len.} & \textbf{Reduc.}
& \textbf{Acc.} & \textbf{Len.} & \textbf{Reduc.}
& \textbf{Acc.} & \textbf{Len.} & \textbf{Reduc.}
& \textbf{Acc.} & \textbf{Len.} & \textbf{Reduc.}
& \textbf{Acc.} & \textbf{Len.} & \textbf{Reduc.} \\
\midrule

% No-controller baseline
\multicolumn{2}{c|}{Vanilla (No Controller)}
& 85.2 & 2857 & 
& 73.5 & 7279 & 
& 50.0 & 10570 & 
& 48.6 & 8347 & 
& 32.8 & 5349 &  \\

\midrule

\multirow{2}{*}{Intervention Position}
& At Transition Words
& 71.2 & \textbf{999} & -65.0\%
& 47.7 & \textbf{1236} & -83.0\%
& 22.7 & \textbf{2828} & -73.2\%
& 32.4 & \textbf{1315} & -84.2\%
& 35.5 & \textbf{1069} & -80.0\% \\

& Every 300 Tokens
& 86.0 & 3638 & +27.3\%
& 68.0 & 9351 & +28.5\%
& 41.3 & 15500 & +46.6\%
& 50.7 & 9587 & +14.9\%
& 29.0 & 15362 & +187.2\% \\

\midrule

Intervention Action
& Add Check Action
& 87.2 & 2025 & -29.1\%
& 70.8 & 4000 & -45.0\%
& 42.7 & 8793 & -16.8\%
& 51.1 & 3888 & -53.4\%
& 38.2 & 3746 & -30.0\% \\

\midrule

Controller Model
& Initialized from Qwen3-1.7B
& \textbf{88.0} & 2295 & -19.7\%
& 72.5 & 4850 & -33.4\%
& 41.3 & 10179 & -3.7\%
& 51.6 & 4946 & -40.7\%
& 40.9 & 4600 & -14.0\% \\

\midrule

% Full method
\rowcolor{gray!18}
\multicolumn{2}{c|}{\textbf{MetaCtrl}}
& \textbf{88.0} & 1644 & -42.5\%
& \textbf{74.7} & 3036 & -58.3\%
& \textbf{56.7} & 5899 & -44.2\%
& \textbf{52.3} & 2965 & -64.5\%
& \textbf{42.4} & 2234 & -58.2\% \\

\bottomrule
\end{tabular}
}
\end{table*}

\subsection{More Ablation Studies}
\label{app:ablation_studies}

\begin{table*}[t]
\centering
\caption{
End-to-end inference latency under Vanilla Reasoning and MetaCtrl Controlled Reasoning with different GPU memory allocations. All experiments are conducted on NVIDIA H20 GPUs. \textbf{Mem. Frac. (R+C)} denotes the configured static GPU memory fractions for the reasoner ($R$) and controller ($C$), respectively. Vanilla Reasoning uses a single GPU for the reasoner. For MetaCtrl with separate GPUs, the reasoner and controller run on two dedicated GPUs. For MetaCtrl with a shared GPU, the reasoner and controller are colocated on the same GPU.
}
\label{tab:e2e_latency_memory}
\resizebox{\textwidth}{!}{
\begin{tabular}{lcccccc}
\toprule
\textbf{Reasoner}
& \multicolumn{2}{c}{\textbf{MATH500}}
& \multicolumn{2}{c}{\textbf{AMC}}
& \multicolumn{2}{c}{\textbf{GPQA Diamond}} \\
\cmidrule(lr){2-3}
\cmidrule(lr){4-5}
\cmidrule(lr){6-7}

& \textbf{Avg. Time (s)} & \textbf{Mem. Frac. (R+C)}
& \textbf{Avg. Time (s)} & \textbf{Mem. Frac. (R+C)}
& \textbf{Avg. Time (s)} & \textbf{Mem. Frac. (R+C)} \\
\midrule

\multicolumn{7}{c}{
\textbf{\textit{Vanilla Reasoning}}
} \\
\midrule

Qwen3-8B
& 31.62 & $0.90 + 0.00$
& 74.93 & $0.90 + 0.00$
& 81.28 & $0.90 + 0.00$ \\

Qwen3-14B
& 45.18 & $0.90 + 0.00$
& 100.94 & $0.90 + 0.00$
& 95.79 & $0.90 + 0.00$ \\

\midrule
\multicolumn{7}{c}{
\textbf{\textit{MetaCtrl Controlled Reasoning (Two Separate GPUs)}}
} \\
\midrule

Qwen3-8B
& 15.44 & $0.90 + 0.60$
& 43.02 & $0.90 + 0.60$
& 23.96 & $0.90 + 0.60$ \\

Qwen3-14B
& 20.63 & $0.90 + 0.60$
& 39.57 & $0.90 + 0.60$
& 36.74 & $0.90 + 0.60$ \\

\midrule
\multicolumn{7}{c}{
\textbf{\textit{MetaCtrl Controlled Reasoning (Two Separate GPUs)}}
} \\
\midrule

Qwen3-8B
& 16.21 & $0.90 + 0.40$
& 38.39 & $0.90 + 0.40$
& 23.77 & $0.90 + 0.40$ \\

Qwen3-14B
& 20.98 & $0.90 + 0.40$
& 38.79 & $0.90 + 0.40$
& 36.95 & $0.90 + 0.40$ \\

\midrule
\multicolumn{7}{c}{
\textbf{\textit{MetaCtrl Controlled Reasoning (Two Separate GPUs)}}
} \\
\midrule

Qwen3-8B
& 15.71 & $0.90 + 0.18$
& 37.32 & $0.90 + 0.18$
& 24.25 & $0.90 + 0.18$ \\

Qwen3-14B
& 21.21 & $0.90 + 0.18$
& 46.93 & $0.90 + 0.18$
& 39.69 & $0.90 + 0.18$ \\

\midrule
\multicolumn{7}{c}{
\textbf{\textit{MetaCtrl Controlled Reasoning (One Shared GPU)}}
} \\
\midrule

Qwen3-8B
& 15.95 & $0.72 + 0.18$
& 37.83 & $0.72 + 0.18$
& 21.74 & $0.72 + 0.18$ \\

Qwen3-14B
& 20.79 & $0.72 + 0.18$
& 48.16 & $0.72 + 0.18$
& 33.33 & $0.72 + 0.18$ \\

\bottomrule
\end{tabular}
}
\end{table*}

\textbf{Controller design.}
Table~\ref{tab:controller_ablation} examines the effects of intervention timing, action space, and controller initialization. Heuristic intervention schedules lead to substantially different tradeoffs. Intervening at transition words produces aggressive compression of 65.0\% to 84.2\%, but substantially reduces accuracy on most benchmarks, whereas intervening every 300 tokens increases generation length on all five benchmarks and underperforms MetaCtrl in accuracy. These results support conditioning intervention timing on the evolving reasoning state rather than using fixed heuristics. Adding a \texttt{Check} action also degrades both accuracy and length reduction relative to the original action space across all benchmarks, suggesting that a larger action space does not necessarily improve control. Similarly, initialization from Qwen3-1.7B results in weaker accuracy and substantially less compression than the default controller on most benchmarks. Overall, the full MetaCtrl design provides the most consistent balance between accuracy and generation length, supporting adaptive intervention timing, a compact action space, and the chosen controller initialization.

\section{The Analysis of Latency and Computation Cost of Methods}
\label{app:latency}

\textbf{End to end latency.} Table~\ref{tab:e2e_latency_memory} shows that MetaCtrl consistently reduces end to end inference latency across both reasoners and all three benchmarks. When the reasoner and controller are placed on two separate GPUs, MetaCtrl reduces mean latency by \textbf{42.6\% to 70.8\%} relative to Vanilla Reasoning across the evaluated memory allocations. This indicates that the reduction in reasoner computation is large enough to offset the additional controller inference and communication overhead. Moreover, reducing the controller memory fraction from 0.60 to 0.18 largely preserves the latency improvement, suggesting that MetaCtrl does not require a heavily provisioned controller to achieve substantial acceleration. The shared GPU setting provides further evidence of practical efficiency. When the reasoner and controller share a single H20 with memory fractions of 0.72 and 0.18, respectively, MetaCtrl reduces latency by \textbf{49.5\% to 73.3\%} across the six reasoner and benchmark combinations, with an average reduction of \textbf{57.3\%}. On GPQA Diamond, for example, latency decreases from 81.28s to 21.74s for Qwen3 8B and from 95.79s to 33.33s for Qwen3 14B. These results show that reductions in reasoning computation translate into substantial wall clock speedups, including when the reasoner and controller are colocated on a single GPU. MetaCtrl does introduce an additional controller model, which increases memory requirements and system complexity compared with Vanilla Reasoning. Nevertheless, the shared GPU results show that this overhead can be reduced substantially while retaining large latency gains.

\section{Training Dynamics and Efficiency
}
\label{app:training_dynamics}

\begin{figure}[t]
    \centering
    \begin{subfigure}[t]{0.48\linewidth}
        \centering
        \includegraphics[width=\linewidth]{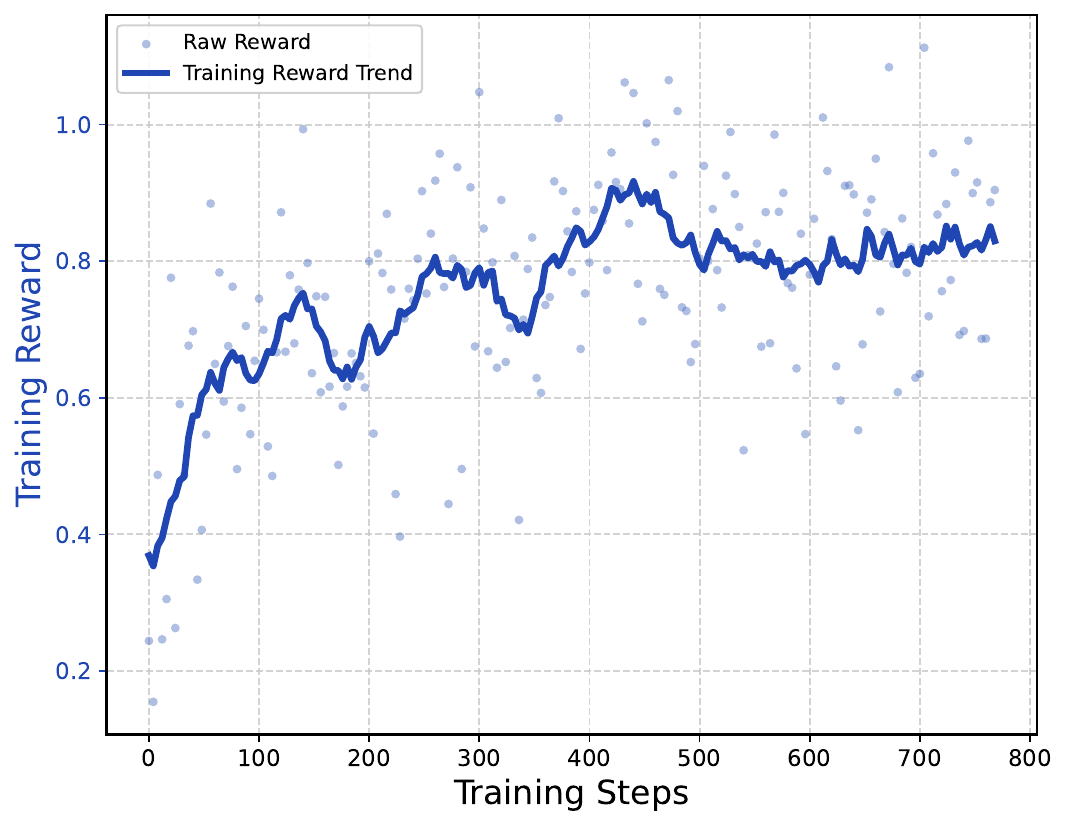}
        \caption{Training reward throughout GRPO optimization.}
        \label{fig:training_reward}
    \end{subfigure}
    \hfill
    \begin{subfigure}[t]{0.48\linewidth}
        \centering
        \includegraphics[width=\linewidth]{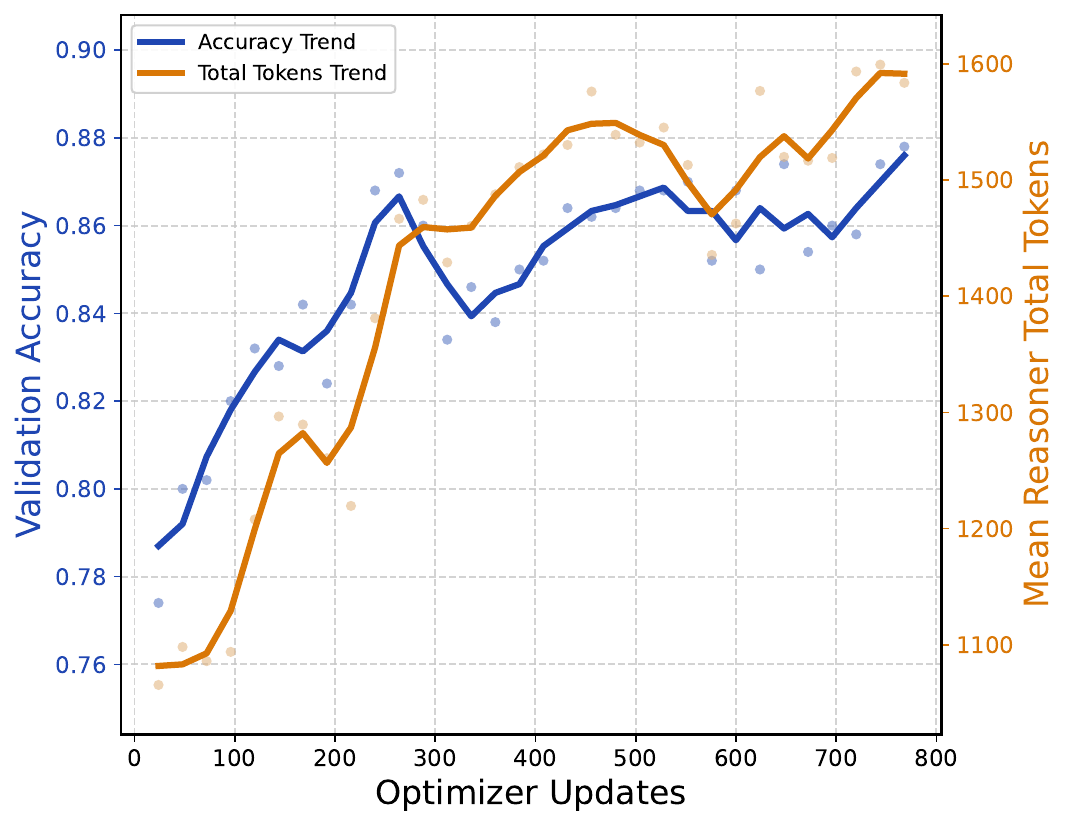}
        \caption{Validation accuracy and average total generation tokens during training.}
        \label{fig:accuracy_tokens}
    \end{subfigure}

    \caption{
    \textbf{Training dynamics of the MetaCtrl.}
    (a) Training reward during GRPO optimization.
    (b) Validation accuracy and average total generation tokens throughout training.
    The reward and accuracy progressively improve and stabilize, while generation length converges to a relatively stable range, indicating that the controller learns to regulate reasoning computation without uniformly compressing reasoning trajectories.
    }
    \label{fig:training_dynamics}
\end{figure}

\paragraph{Stable policy improvement during training.} We analyze the optimization dynamics of our MetaCtrl by tracking the training reward throughout GRPO training (Figure~\ref{fig:training_reward}). The reward increases rapidly during the early stage of training and continues to improve despite local fluctuations, reaching a substantially higher level after roughly 400 training steps. It subsequently stabilizes within a relatively narrow range, indicating that the controller progressively learns more effective intervention policies rather than relying on transient high-reward behaviors. The sustained improvement and stable late-stage reward further suggest that the proposed optimization objective provides a reliable learning signal for regulating the reasoning process.

\begin{figure}[t]
	\centering
	\includegraphics[width=\linewidth]{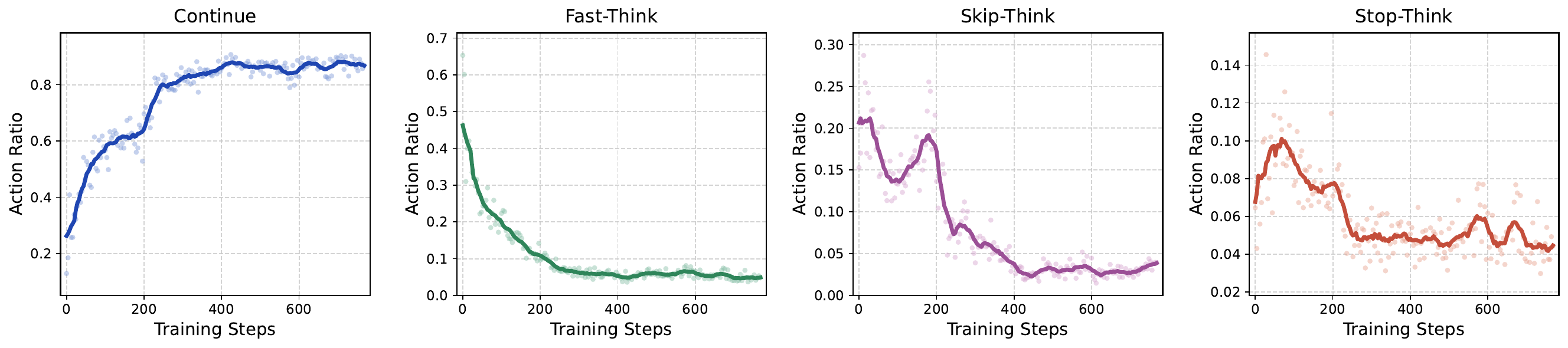}
	\caption{
    \textbf{Evolution of the controller action distribution during training.}
    The learned policy gradually shifts toward \textsc{Continue}, while the frequencies of \textsc{Fast-Think}, \textsc{Skip-Think}, and \textsc{Stop-Think} decrease and stabilize at lower levels. This trend suggests that the controller progressively learns to intervene more selectively rather than frequently modifying the reasoner's trajectory.
    }
    \label{fig:controller_action_distribution_trend}
\end{figure}

\paragraph{Balancing accuracy and reasoning length.}
We track validation accuracy and mean reasoner generation length throughout training in Figure~\ref{fig:accuracy_tokens}. Validation accuracy increases from approximately \(79\%\) at the beginning of training to around \(87\%\) in later stages, while mean reasoning length rises from roughly \(1.1\)K to around \(1.55\)K tokens. The two metrics exhibit broadly aligned trends during optimization, indicating that training does not collapse to a trivial policy that simply minimizes reasoning length. Instead, the controller learns to retain additional computation when it is useful for correctness, rather than aggressively compressing reasoning at the expense of accuracy.

\paragraph{Evolution of the controller policy.}
We further investigate how the learned control strategy evolves during training by tracking the distribution of the four controller actions in Figure~\ref{fig:controller_action_distribution_trend}. At the beginning of training, the controller frequently modifies the reasoner's trajectory through \textsc{Fast-Think}, \textsc{Skip-Think}, and \textsc{Stop-Think}, while \textsc{Continue} constitutes only a relatively small fraction of the decisions. As optimization proceeds, the action distribution changes markedly: the proportion of \textsc{Continue} steadily increases and eventually dominates the policy, whereas \textsc{Fast-Think} and \textsc{Skip-Think} decrease substantially and \textsc{Stop-Think} remains comparatively infrequent. Notably, the intervention actions do not vanish completely, but converge to small yet non-zero frequencies. This behavior suggests that the controller does not learn to indiscriminately shorten the frozen reasoner's reasoning process. Instead, it progressively learns to preserve the reasoner's native trajectory in most cases and intervene selectively when modifying the ongoing reasoning is useful. Together with the improvements in reward and accuracy and the stabilization of generation length, these dynamics indicate the emergence of a stable and selective computation policy: the controller learns not only how to intervene, but also when intervention is unnecessary.

\section{Future work}
A promising direction is to extend MetaCtrl to a collaborative multi-reasoner system, where multiple reasoning models constitute the object-level and a shared meta-level controller coordinates the overall reasoning process. For each problem, the controller could first perform joint model routing and initial effort allocation, selecting which reasoner should begin solving the problem and under which reasoning effort configuration.

As reasoning unfolds, the controller could continuously monitor the evolving explicit reasoning trace and decide whether to let the current reasoner proceed, intervene through lightweight textual guidance to regulate its subsequent reasoning, or selectively invoke a stronger model for targeted guidance, verification, or correction. Routine reasoning could therefore remain with smaller and less expensive models, while stronger models are reserved for stages where additional capability is most valuable. The resulting system would jointly control computation at two levels: across models, by deciding which reasoner should handle each stage of the solution process, and within a selected reasoner, by regulating how its ongoing reasoning should proceed through trajectory-level interventions.

This setting raises an important resource allocation problem: when a difficult intermediate state should be handled by further regulating the current reasoner, and when it should instead be escalated to a more capable model, selectively invoked at critical stages to provide targeted guidance or correction. Learning this trade-off online could reduce the need for users to manually choose among model families and reasoning-effort settings, while avoiding the cost of running the strongest model throughout the entire solution process. More broadly, such a system would extend metacognitive control from regulating a single reasoning trajectory to coordinating capability and computation across a collaborative reasoning system, with the goal of achieving a desired level of reliability at lower end-to-end inference cost.

\section{Case Study}
\label{app:case_study}
\subsection{Preventing Post-Solution Overthinking}

\begin{figure}[t]
	\centering
	\includegraphics[width=\linewidth]{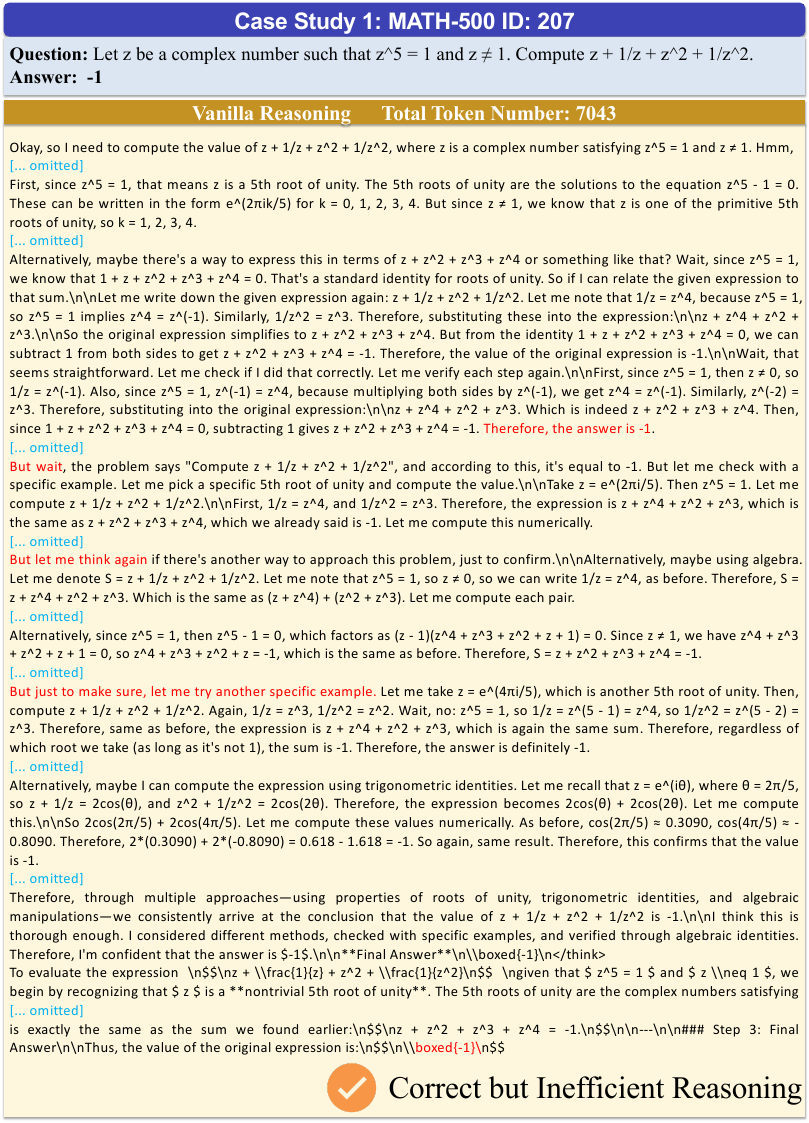}
	\caption{Vanilla reasoning on a MATH-500 example. The model obtains the correct answer but continues with repetitive verification and alternative derivations, leading to substantial overthinking and 7,043 reasoning tokens.}
    \label{fig:case_study1}
\end{figure}

\begin{figure}[t]
	\centering
	\includegraphics[width=\linewidth]{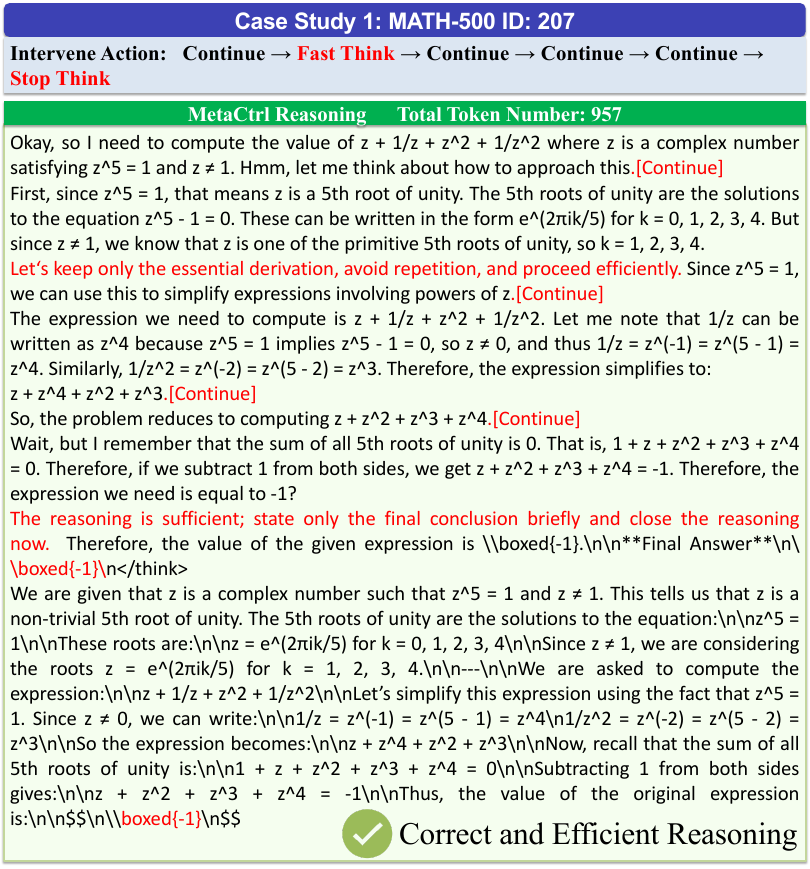}
	\caption{MetaCtrl reasoning on the MATH-500 example. Both methods arrive at the correct answer, while MetaCtrl dynamically regulates the reasoning trajectory through lightweight interventions and terminates once the solution is sufficiently established, reducing the reasoning length from 7,043 to 957 tokens.}
    \label{fig:case_study2}
\end{figure}

\begin{figure}[t]
	\centering
	\includegraphics[width=\linewidth]{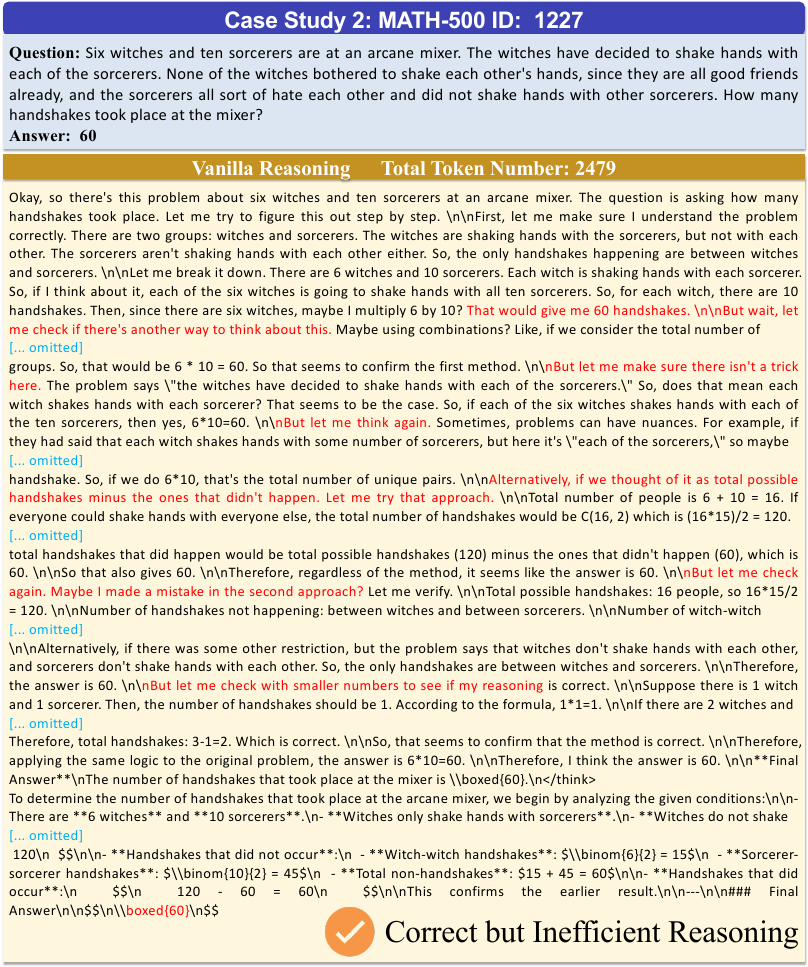}
	\caption{Vanilla reasoning on a MATH-500 example. The model reaches the correct answer early but continues with alternative counting strategies, repeated verification, and additional sanity checks, resulting in 2,479 reasoning tokens.}
    \label{fig:case_study3}
\end{figure}

\begin{figure}[t]
	\centering
	\includegraphics[width=\linewidth]{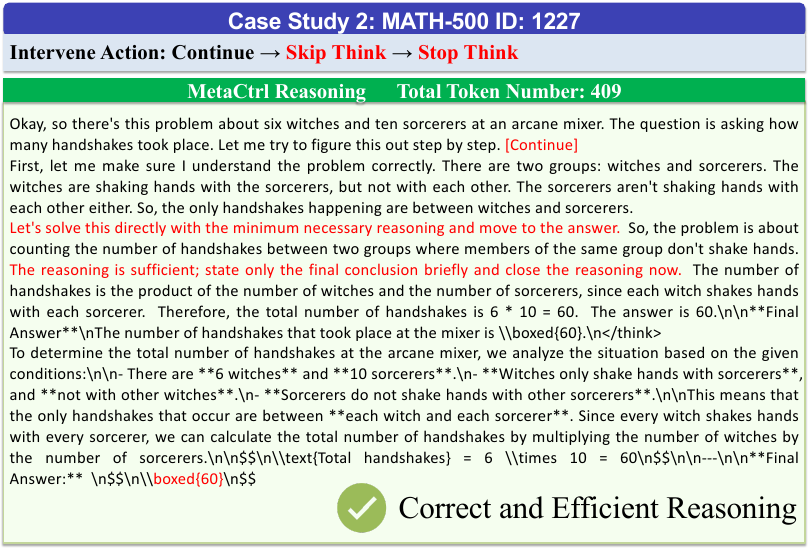}
	\caption{MetaCtrl reasoning on the MATH-500 example. After identifying that all valid handshakes occur between the two groups, the controller skips unnecessary intermediate reasoning and terminates once sufficient information for the solution has been established, reducing the reasoning length from 2,479 to 409 tokens while preserving the correct answer.}
    \label{fig:case_study4}
\end{figure}

\begin{figure}[t]
	\centering
	\includegraphics[width=\linewidth]{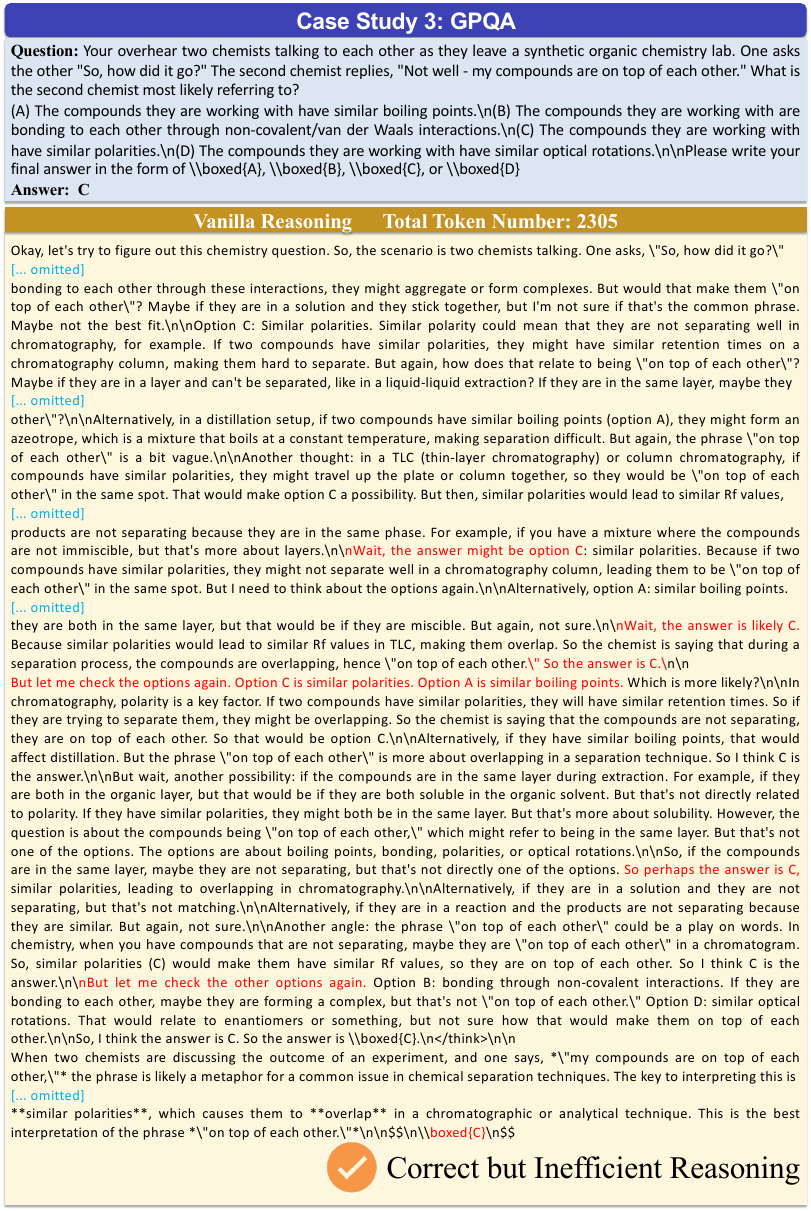}
	\caption{Vanilla reasoning on a GPQA example. Although the model eventually selects the correct answer, it repeatedly revisits competing explanations and re-evaluates previously considered options, resulting in 2,305 reasoning tokens.}
    \label{fig:case_study5}
\end{figure}

\begin{figure}[t]
	\centering
	\includegraphics[width=\linewidth]{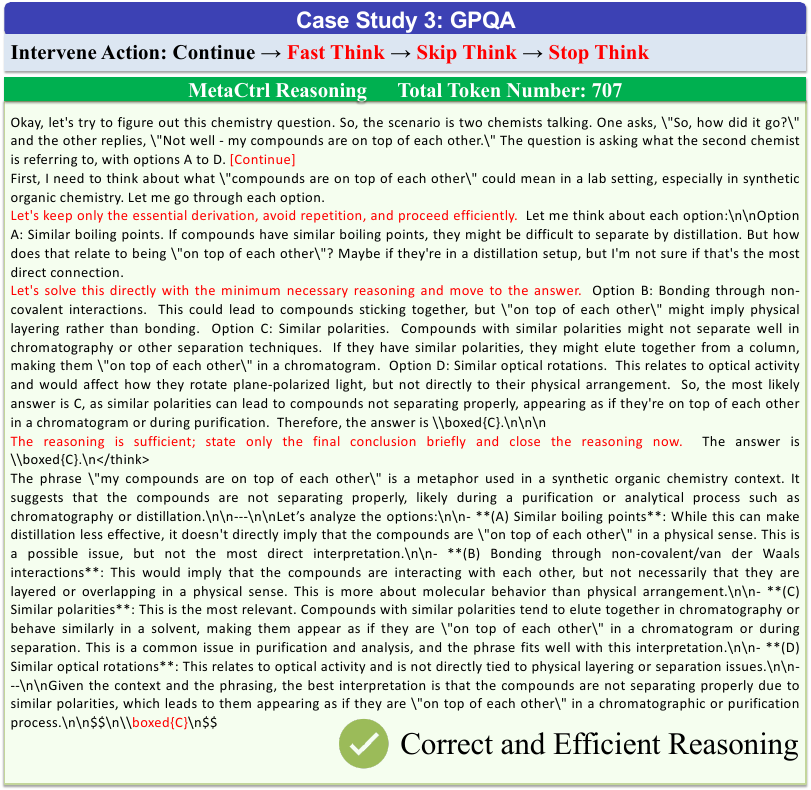}
	\caption{MetaCtrl reasoning on the same GPQA example. The controller progressively regulates the reasoning trajectory through Fast Think, Skip Think, and Stop Think interventions, reducing repeated hypothesis exploration and reaching the same correct answer with 707 reasoning tokens.}
    \label{fig:case_study6}
\end{figure}

\begin{figure}[t]
	\centering
	\includegraphics[width=0.95\linewidth]{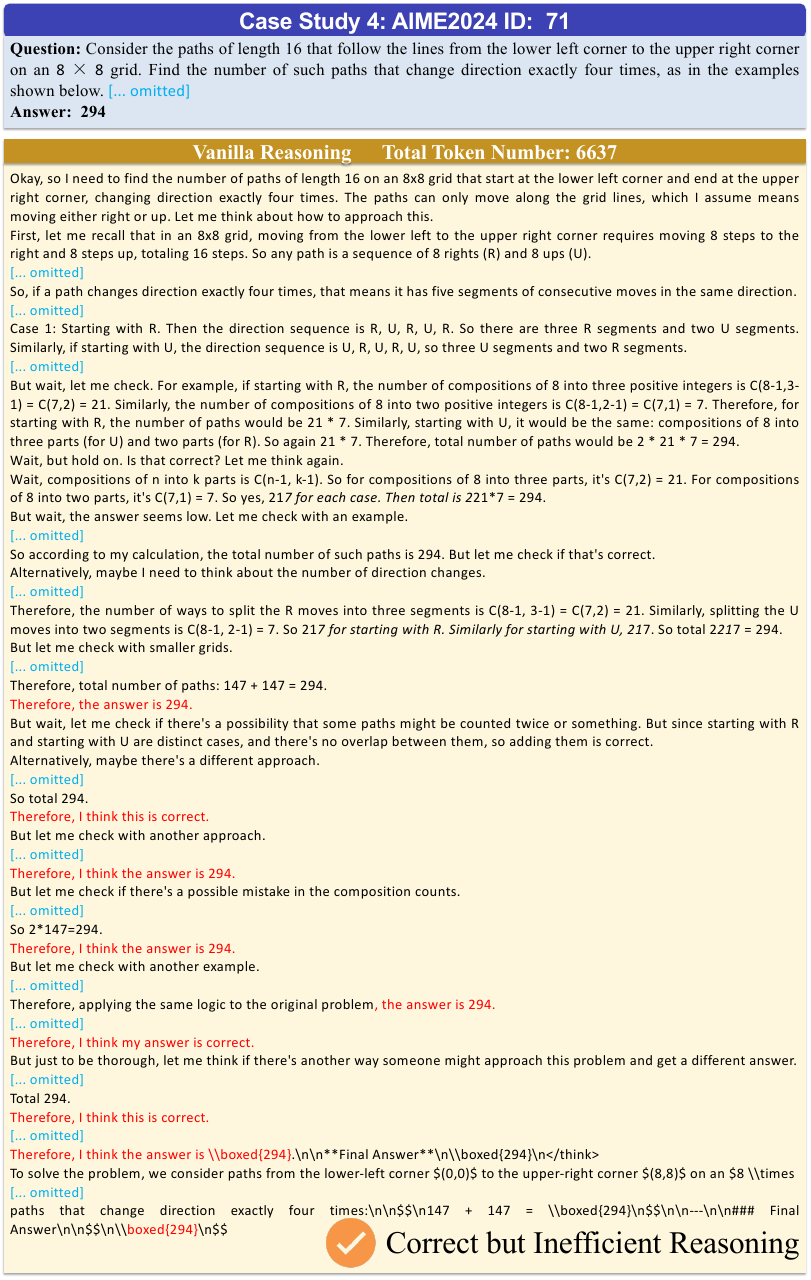}
	\caption{Vanilla reasoning on an AIME 2024 example. The model derives the correct composition-based counting argument but continues to revisit the same calculation through repeated verification and alternative checks, resulting in 6,637 reasoning tokens.}
    \label{fig:case_study7}
\end{figure}

\begin{figure}[t]
	\centering
	\includegraphics[width=\linewidth]{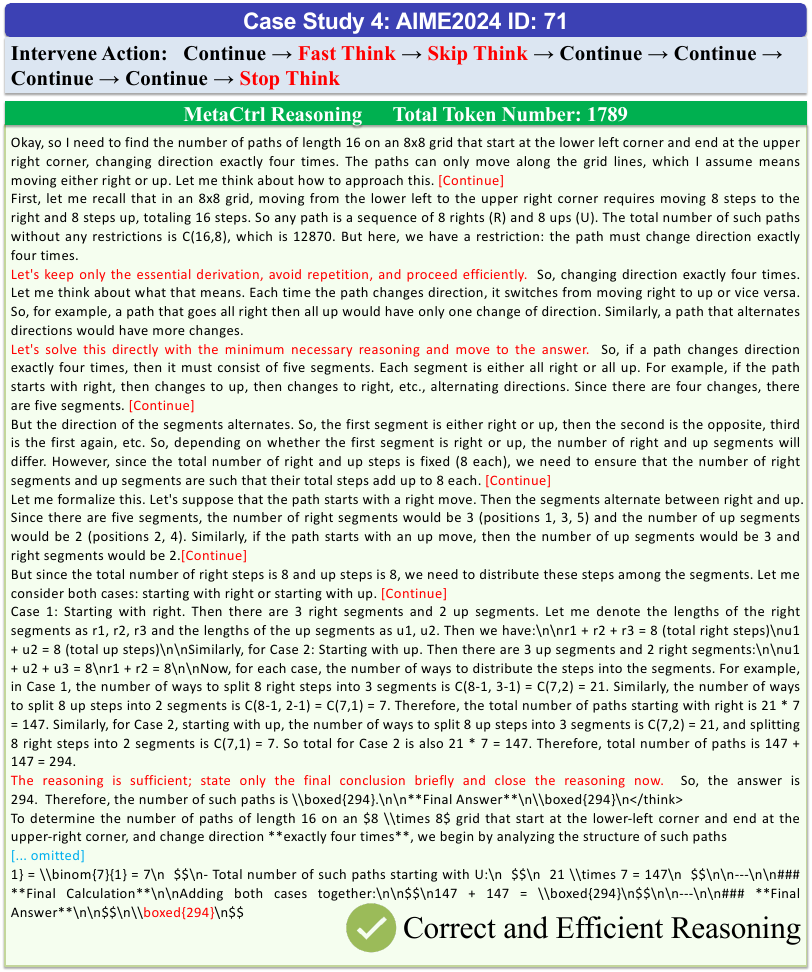}
	\caption{MetaCtrl reasoning on the same AIME 2024 example. The controller dynamically accelerates the reasoning trajectory and terminates once the composition-based counting argument is complete, reaching the same correct answer with 1,789 reasoning tokens.}
    \label{fig:case_study8}
\end{figure}

\begin{figure}[t]
	\centering
	\includegraphics[width=\linewidth]{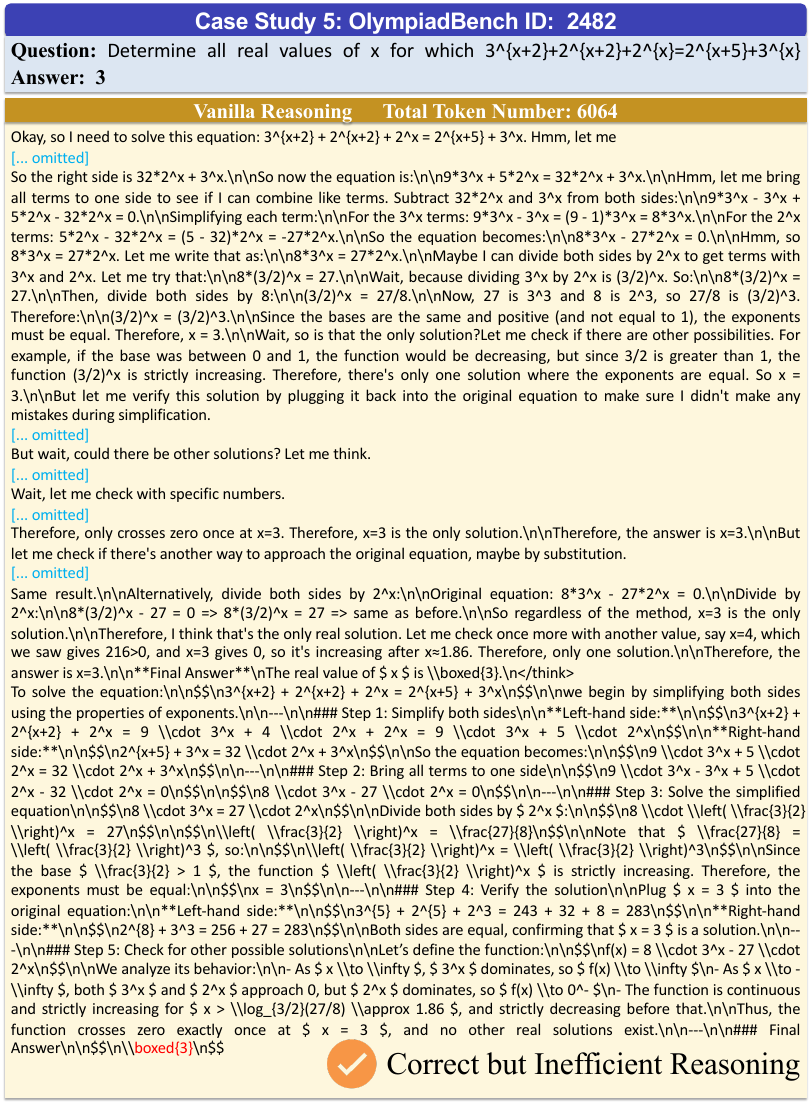}
	\caption{Vanilla reasoning on an OlympiadBench example. The model correctly derives the unique solution but continues with repeated substitution checks, alternative derivations, and unnecessary uniqueness verification, resulting in 6,064 reasoning tokens.}
    \label{fig:case_study9}
\end{figure}

\begin{figure}[t]
	\centering
	\includegraphics[width=\linewidth]{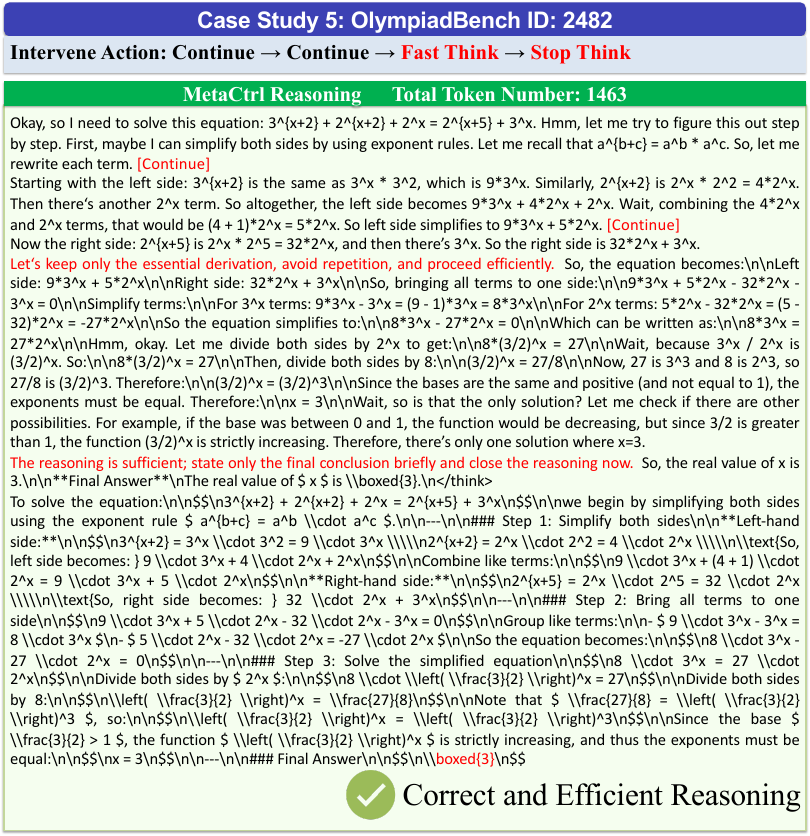}
	\caption{MetaCtrl reasoning on the same OlympiadBench example. After the key algebraic reduction establishes the unique solution, the controller terminates further deliberation, reaching the same correct answer with 1,463 reasoning tokens.}
    \label{fig:case_study10}
\end{figure}

As shown in Figures~\ref{fig:case_study1}--\ref{fig:case_study10}, across five representative reasoning problems, vanilla reasoning often continues with redundant verification or repeated hypothesis exploration after the key solution has already been established. In contrast, MetaCtrl adaptively regulates the evolving trajectory, preserving necessary derivations while accelerating or terminating further computation when sufficient evidence has accumulated. The consistent preservation of correct answers across these cases illustrates how MetaCtrl reduces diverse forms of overthinking without aggressively truncating reasoning.

\subsection{Reducing Errors from Prolonged Reasoning}

\begin{figure}[t]
	\centering
	\includegraphics[width=\linewidth]{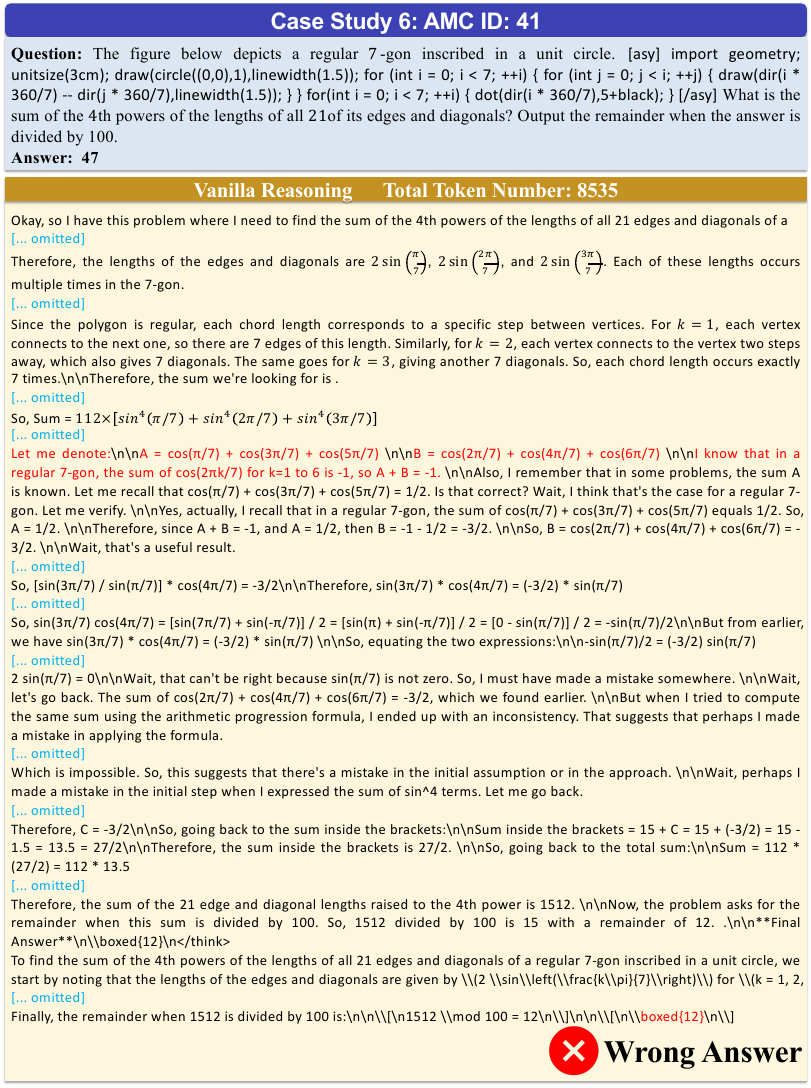}
	\caption{Vanilla reasoning on an AMC example. The model enters an inconsistent trigonometric detour and continues reasoning despite recognizing contradictions in its intermediate derivation, ultimately producing an incorrect answer after 8,535 reasoning tokens.}
    \label{fig:case_study11}
\end{figure}

\begin{figure}[t]
	\centering
	\includegraphics[width=\linewidth]{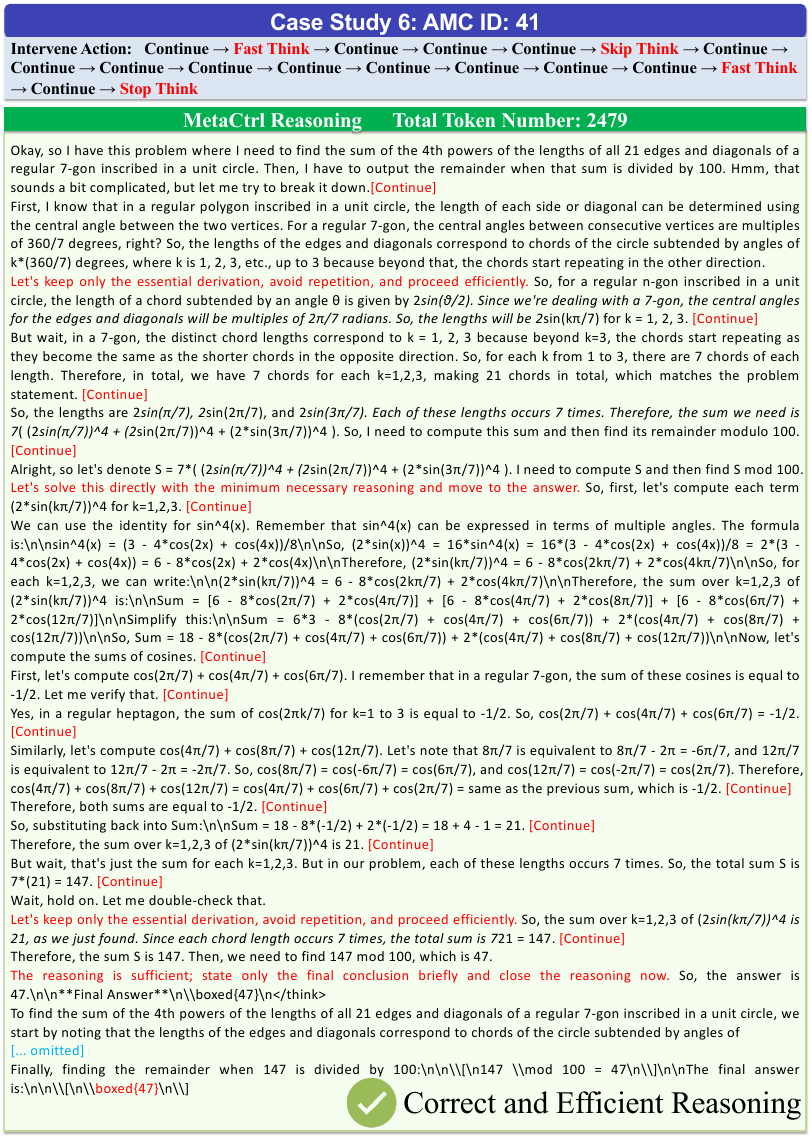}
	\caption{MetaCtrl reasoning on the same AMC example. By dynamically regulating how the frozen reasoner proceeds, MetaCtrl follows a more direct derivation and avoids prolonged unproductive deliberation, reaching the correct answer with 2,479 reasoning tokens.}
    \label{fig:case_study12}
\end{figure}

\begin{figure}[t]
	\centering
	\includegraphics[width=0.95\linewidth]{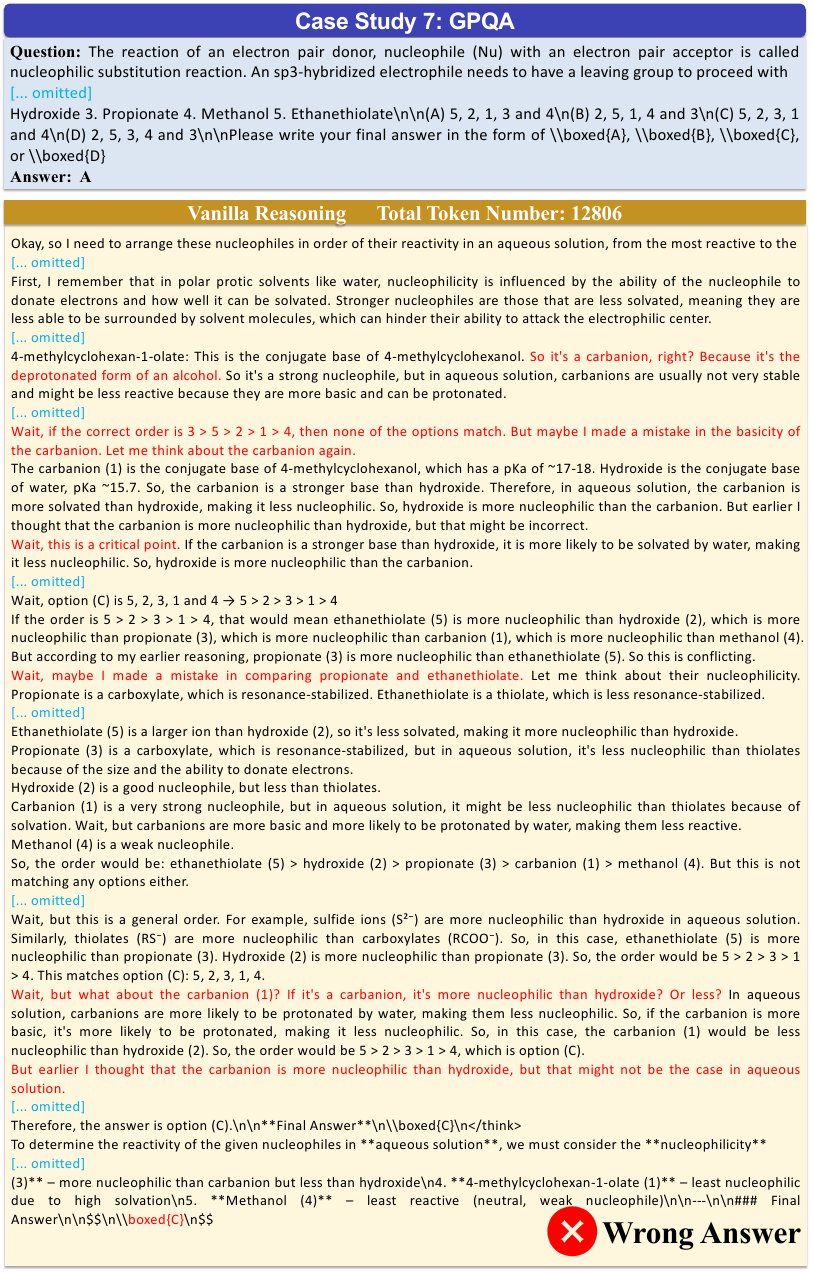}
	\caption{Vanilla reasoning on a GPQA chemistry example. The model repeatedly revisits conflicting hypotheses about solvation, basicity, and nucleophilicity, and an early mischaracterization of the alkoxide contributes to a prolonged inconsistent trajectory, ultimately yielding an incorrect answer after 12,806 reasoning tokens.}
    \label{fig:case_study13}
\end{figure}

\begin{figure}[t]
	\centering
	\includegraphics[width=\linewidth]{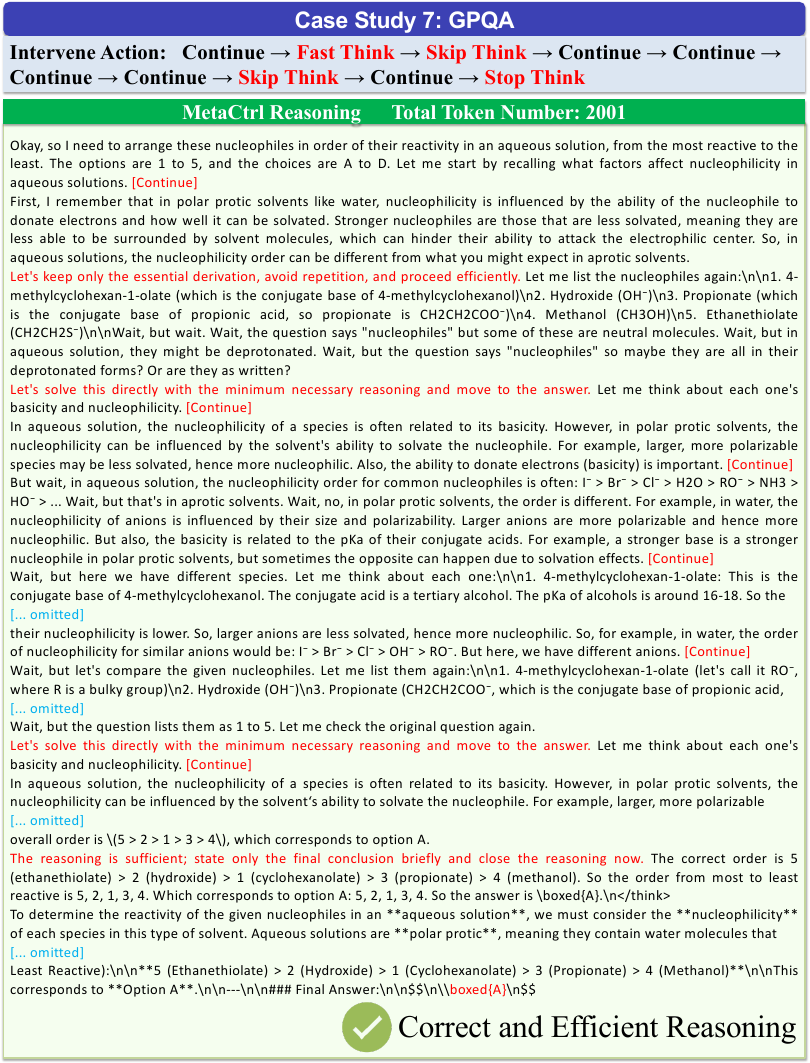}
	\caption{MetaCtrl reasoning on the same GPQA example. By dynamically regulating how the frozen reasoner proceeds, MetaCtrl substantially reduces prolonged hypothesis exploration and reaches the correct answer with 2,001 reasoning tokens.}
    \label{fig:case_study14}
\end{figure}

\begin{figure}[t]
	\centering
	\includegraphics[width=0.95\linewidth]{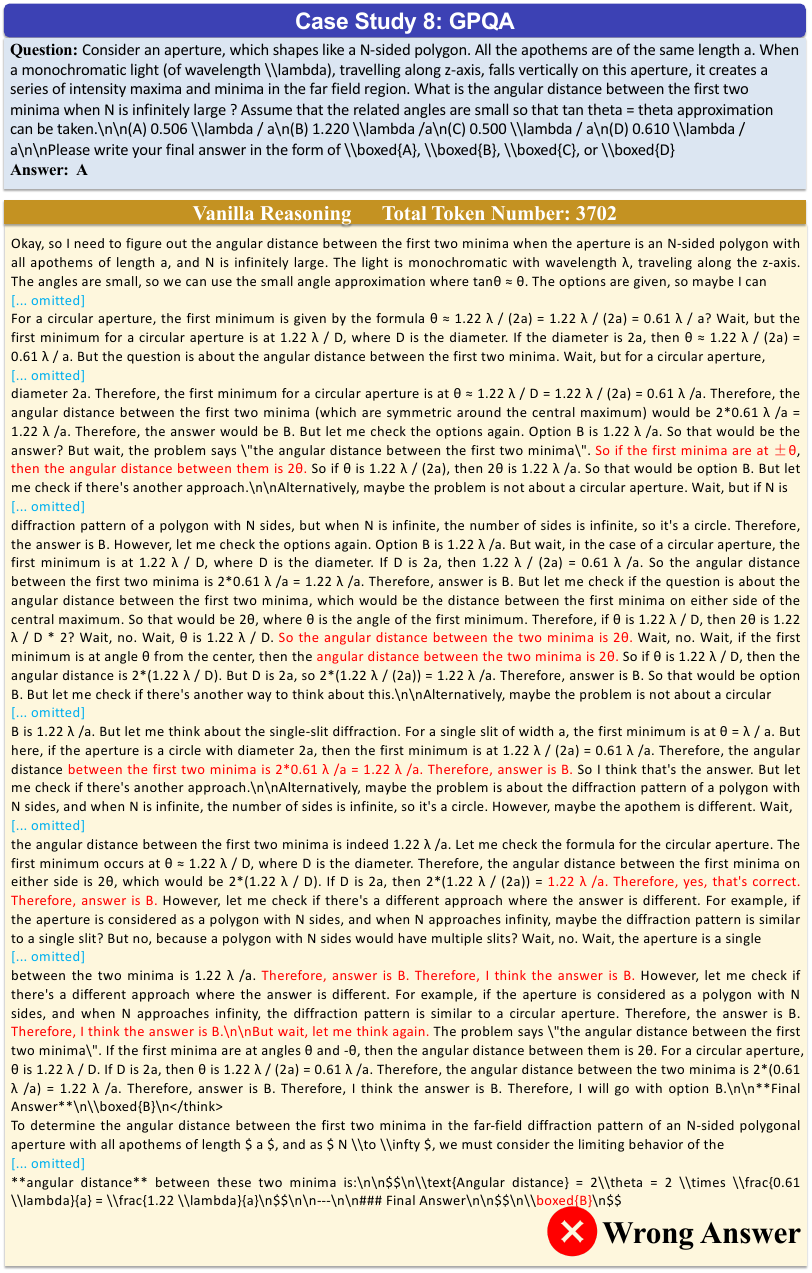}
	\caption{Vanilla reasoning on a GPQA physics example. The model repeatedly interprets the “first two minima” as the symmetric first minima on opposite sides of the central maximum, leading to prolonged verification of the same interpretation and ultimately an incorrect answer after 3,702 reasoning tokens.}
    \label{fig:case_study15}
\end{figure}

\begin{figure}[t]
	\centering
	\includegraphics[width=0.9\linewidth]{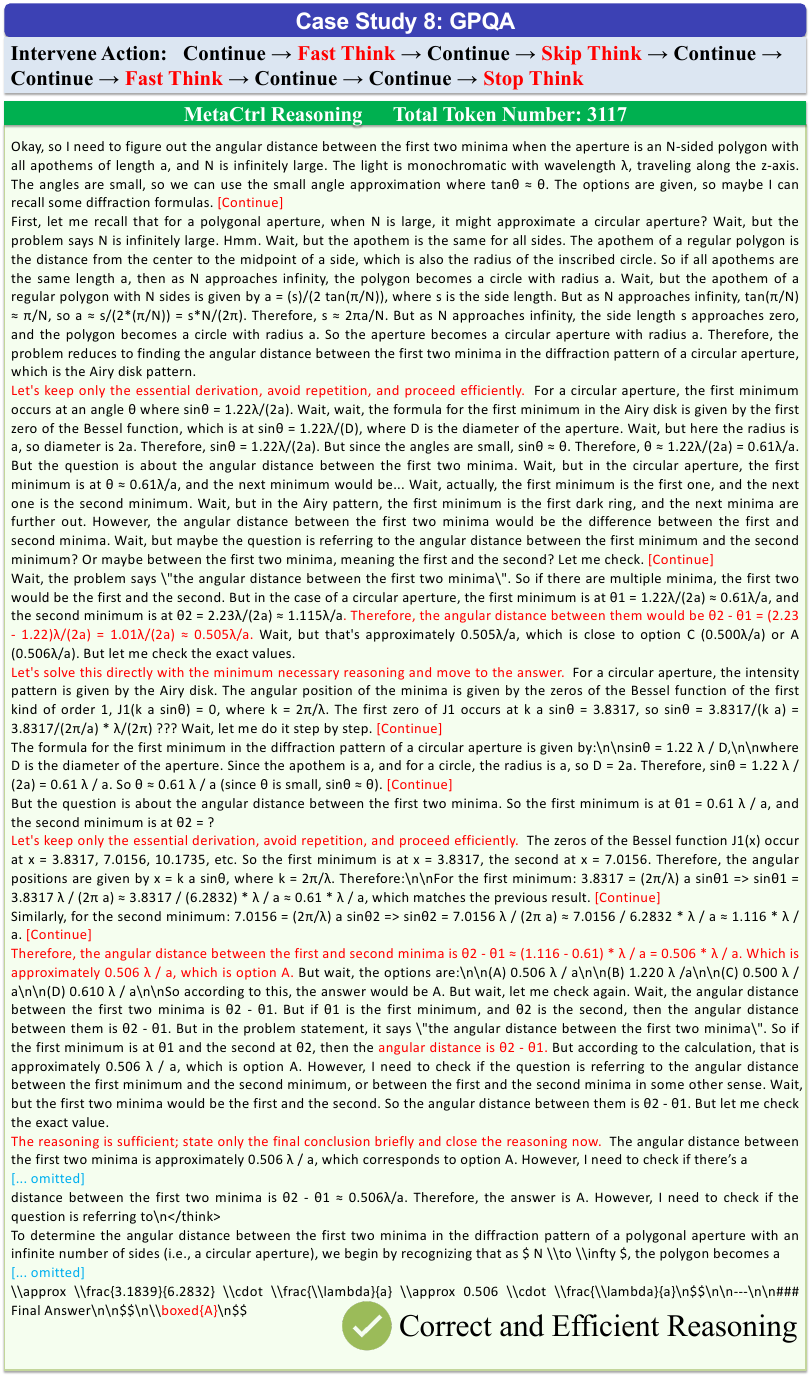}
	\caption{MetaCtrl reasoning on the same GPQA example. The controlled trajectory revisits the interpretation of the diffraction minima and uses the first two zeros of the Airy pattern, reaching the correct answer with 3,117 reasoning tokens.}
    \label{fig:case_study16}
\end{figure}

As shown in Figures~\ref{fig:case_study11}--\ref{fig:case_study16}, these cases further illustrate that longer reasoning is not necessarily more reliable. Vanilla reasoning can become trapped in prolonged hypothesis exploration or increasingly complex derivations, where additional deliberation may compound inconsistencies and ultimately lead to incorrect answers. In contrast, MetaCtrl regulates how the frozen reasoner proceeds through lightweight interventions, encouraging more direct reasoning and terminating unnecessary deliberation once a sufficient solution path has emerged. In these examples, such controlled trajectories are not only substantially shorter but also reach the correct answers, suggesting that simpler and more focused reasoning can sometimes be more reliable than unconstrained overthinking.

\end{document}

%% file: math_commands.tex
\usepackage{amsmath,amsfonts,bm}

\def\eqref#1{equation~\ref{#1}}
\def\1{\bm{1}}

\DeclareMathAlphabet{\mathsfit}{\encodingdefault}{\sfdefault}{m}{sl}
\SetMathAlphabet{\mathsfit}{bold}{\encodingdefault}{\sfdefault}{bx}{n}

%% file: iclr2027_conference.bbl
\begin{thebibliography}{75}
\providecommand{\natexlab}[1]{#1}
\providecommand{\url}[1]{\texttt{#1}}
\expandafter\ifx\csname urlstyle\endcsname\relax
  \providecommand{\doi}[1]{doi: #1}\else
  \providecommand{\doi}{doi: \begingroup \urlstyle{rm}\Url}\fi

\bibitem[Ackerman \& Thompson(2017)Ackerman and Thompson]{ackerman2017meta}
Rakefet Ackerman and Valerie~A Thompson.
\newblock Meta-reasoning: Monitoring and control of thinking and reasoning.
\newblock \emph{Trends in cognitive sciences}, 21\penalty0 (8):\penalty0 607--617, 2017.

\bibitem[Aggarwal \& Welleck(2025)Aggarwal and Welleck]{aggarwal2025l1}
Pranjal Aggarwal and Sean Welleck.
\newblock L1: Controlling how long a reasoning model thinks with reinforcement learning.
\newblock \emph{arXiv preprint arXiv:2503.04697}, 2025.

\bibitem[{AI-MO}(2024)]{aimo2024amc}
{AI-MO}.
\newblock {AMC} 2023, 2024.
\newblock URL \url{https://huggingface.co/datasets/AI-MO/aimo-validation-amc}.

\bibitem[Aytes et~al.(2025)Aytes, Baek, and Hwang]{aytes2025sketch}
Simon~A Aytes, Jinheon Baek, and Sung~Ju Hwang.
\newblock Sketch-of-thought: Efficient llm reasoning with adaptive cognitive-inspired sketching.
\newblock In \emph{Proceedings of the 2025 Conference on Empirical Methods in Natural Language Processing}, pp.\  24307--24331, 2025.

\bibitem[Cao et~al.(2025)Cao, Zou, Peng, Chen, Ning, and Li]{cao2025step}
Lang Cao, Yingtian Zou, Chao Peng, Renhong Chen, Wu~Ning, and Yitong Li.
\newblock Step guided reasoning: Improving mathematical reasoning using guidance generation and step reasoning.
\newblock In \emph{Proceedings of the 2025 Conference on Empirical Methods in Natural Language Processing}, pp.\  21112--21129, 2025.

\bibitem[Chen et~al.(2025{\natexlab{a}})Chen, Zhang, Hong, Kundu, and Wang]{chen2025seal}
Runjin Chen, Zhenyu Zhang, Junyuan Hong, Souvik Kundu, and Zhangyang Wang.
\newblock Seal: Steerable reasoning calibration of large language models for free.
\newblock \emph{arXiv preprint arXiv:2504.07986}, 2025{\natexlab{a}}.

\bibitem[Chen et~al.(2026)Chen, Jiang, Zhang, Chen, Yan, Xie, Yang, and Huang]{chen2026reasoning}
Xin Chen, Feng Jiang, Yiqian Zhang, Hardy Chen, Shuo Yan, Wenya Xie, Min Yang, and Shujian Huang.
\newblock Reasoning while asking: Transforming reasoning large language models from passive solvers to proactive inquirers.
\newblock In \emph{Proceedings of the 64th Annual Meeting of the Association for Computational Linguistics (Volume 1: Long Papers)}, pp.\  35069--35090, 2026.

\bibitem[Chen et~al.(2025{\natexlab{b}})Chen, Xu, Liang, He, Pang, Yu, Song, Liu, Zhou, Zhang, et~al.]{chen2025not}
Xingyu Chen, Jiahao Xu, Tian Liang, Zhiwei He, Jianhui Pang, Dian Yu, Linfeng Song, Qiuzhi Liu, Mengfei Zhou, Zhuosheng Zhang, et~al.
\newblock Do not think that much for 2+ 3=? on the overthinking of long reasoning models.
\newblock In \emph{Forty-second International Conference on Machine Learning}, 2025{\natexlab{b}}.

\bibitem[Ding et~al.(2024)Ding, Liu, Fu, Song, Xie, and Zhang]{ding2024break}
Mengru Ding, Hanmeng Liu, Zhizhang Fu, Jian Song, Wenbo Xie, and Yue Zhang.
\newblock Break the chain: Large language models can be shortcut reasoners.
\newblock \emph{arXiv preprint arXiv:2406.06580}, 2024.

\bibitem[Fu et~al.(2025)Fu, Chen, Zhu, Fu, Dai, Zhuang, Ma, Qiao, Rosing, Stoica, and Zhang]{fu2025efficiently}
Yichao Fu, Junda Chen, Siqi Zhu, Zheyu Fu, Zhongdongming Dai, Yonghao Zhuang, Yian Ma, Aurick Qiao, Tajana Rosing, Ion Stoica, and Hao Zhang.
\newblock Efficiently scaling {LLM} reasoning programs with certaindex.
\newblock In \emph{The Thirty-ninth Annual Conference on Neural Information Processing Systems}, 2025.
\newblock URL \url{https://openreview.net/forum?id=nn51ewu5k2}.

\bibitem[Gao et~al.(2025)Gao, Song, Yang, Cai, Miao, Dong, Li, Ma, Chen, Tang, et~al.]{gao2025omni}
Bofei Gao, Feifan Song, Zhe Yang, Zefan Cai, Yibo Miao, Qingxiu Dong, Lei Li, Chenghao Ma, Liang Chen, Zhengyang Tang, et~al.
\newblock Omni-math: A universal olympiad level mathematic benchmark for large language models.
\newblock In \emph{International Conference on Learning Representations}, volume 2025, pp.\  100540--100569, 2025.

\bibitem[Guo et~al.(2025)Guo, Yang, Zhang, Song, Wang, Zhu, Xu, Zhang, Ma, Bi, et~al.]{guo2025deepseek}
Daya Guo, Dejian Yang, Haowei Zhang, Junxiao Song, Peiyi Wang, Qihao Zhu, Runxin Xu, Ruoyu Zhang, Shirong Ma, Xiao Bi, et~al.
\newblock Deepseek-r1 incentivizes reasoning in llms through reinforcement learning.
\newblock \emph{Nature}, 645\penalty0 (8081):\penalty0 633--638, 2025.

\bibitem[Han et~al.(2025)Han, Wang, Fang, Zhao, Ma, and Chen]{han2025token}
Tingxu Han, Zhenting Wang, Chunrong Fang, Shiyu Zhao, Shiqing Ma, and Zhenyu Chen.
\newblock Token-budget-aware llm reasoning.
\newblock In \emph{Findings of the Association for Computational Linguistics: ACL 2025}, pp.\  24842--24855, 2025.

\bibitem[Hao et~al.(2024)Hao, Sukhbaatar, Su, Li, Hu, Weston, and Tian]{hao2024training}
Shibo Hao, Sainbayar Sukhbaatar, DiJia Su, Xian Li, Zhiting Hu, Jason Weston, and Yuandong Tian.
\newblock Training large language models to reason in a continuous latent space.
\newblock \emph{arXiv preprint arXiv:2412.06769}, 2024.

\bibitem[He et~al.(2024)He, Luo, Bai, Hu, Thai, Shen, Hu, Han, Huang, Zhang, et~al.]{he2024olympiadbench}
Chaoqun He, Renjie Luo, Yuzhuo Bai, Shengding Hu, Zhen Thai, Junhao Shen, Jinyi Hu, Xu~Han, Yujie Huang, Yuxiang Zhang, et~al.
\newblock Olympiadbench: A challenging benchmark for promoting agi with olympiad-level bilingual multimodal scientific problems.
\newblock In \emph{Proceedings of the 62nd Annual Meeting of the Association for Computational Linguistics (Volume 1: Long Papers)}, pp.\  3828--3850, 2024.

\bibitem[Hendrycks et~al.(2021)Hendrycks, Burns, Kadavath, Arora, Basart, Tang, Song, and Steinhardt]{hendrycks2021measuring}
Dan Hendrycks, Collin Burns, Saurav Kadavath, Akul Arora, Steven Basart, Eric Tang, Dawn Song, and Jacob Steinhardt.
\newblock Measuring mathematical problem solving with the math dataset.
\newblock \emph{arXiv preprint arXiv:2103.03874}, 2021.

\bibitem[Hou et~al.(2025)Hou, Zhang, Ji, Liu, Qian, Andreas, and Chang]{hou2025thinkprune}
Bairu Hou, Yang Zhang, Jiabao Ji, Yujian Liu, Kaizhi Qian, Jacob Andreas, and Shiyu Chang.
\newblock Thinkprune: Pruning long chain-of-thought of llms via reinforcement learning.
\newblock \emph{arXiv preprint arXiv:2504.01296}, 2025.

\bibitem[Huang et~al.(2026)Huang, Lin, Feng, Chen, He, and Hou]{huang2026efficient}
Jiameng Huang, Baijiong Lin, Guhao Feng, Jierun Chen, Di~He, and Lu~Hou.
\newblock Efficient reasoning for large reasoning language models via certainty-guided reflection suppression.
\newblock In \emph{Proceedings of the AAAI Conference on Artificial Intelligence}, volume~40, pp.\  31176--31184, 2026.

\bibitem[Huang et~al.(2025)Huang, Wang, Zhong, Su, Feng, Cao, and Fung]{huang2025adactrl}
Shijue Huang, Hongru Wang, Wanjun Zhong, Zhaochen Su, Jiazhan Feng, Bowen Cao, and Yi~R Fung.
\newblock Adactrl: Towards adaptive and controllable reasoning via difficulty-aware budgeting.
\newblock \emph{arXiv preprint arXiv:2505.18822}, 2025.

\bibitem[{Hugging Face}(2025)]{openr1}
{Hugging Face}.
\newblock Open r1: A fully open reproduction of deepseek-r1, January 2025.
\newblock URL \url{https://github.com/huggingface/open-r1}.

\bibitem[Jaech et~al.(2024)Jaech, Kalai, Lerer, Richardson, El-Kishky, Low, Helyar, Madry, Beutel, Carney, et~al.]{jaech2024openai}
Aaron Jaech, Adam Kalai, Adam Lerer, Adam Richardson, Ahmed El-Kishky, Aiden Low, Alec Helyar, Aleksander Madry, Alex Beutel, Alex Carney, et~al.
\newblock Openai o1 system card.
\newblock \emph{arXiv preprint arXiv:2412.16720}, 2024.

\bibitem[Jain et~al.(2025)Jain, Han, Gu, Li, Yan, Zhang, Wang, Solar-Lezama, Sen, and Stoica]{jain2025livecodebench}
Naman Jain, King Han, Alex Gu, Wen-Ding Li, Fanjia Yan, Tianjun Zhang, Sida Wang, Armando Solar-Lezama, Koushik Sen, and Ion Stoica.
\newblock Livecodebench: Holistic and contamination free evaluation of large language models for code.
\newblock In \emph{International Conference on Learning Representations}, volume 2025, pp.\  58791--58831, 2025.

\bibitem[Jia et~al.(2026)Jia, Zhang, Diao, Yuan, Ouyang, Ma, and Vosoughi]{jia2026makes}
Yaning Jia, Chunhui Zhang, Xingjian Diao, Xiangchi Yuan, Zhongyu Ouyang, Chiyu Ma, and Soroush Vosoughi.
\newblock What makes a good curriculum? disentangling the effects of data ordering on llm mathematical reasoning.
\newblock In \emph{Proceedings of the 64th Annual Meeting of the Association for Computational Linguistics (Volume 1: Long Papers)}, pp.\  34472--34488, 2026.

\bibitem[Jin et~al.(2024)Jin, Yu, Shu, Zhao, Hua, Meng, Zhang, and Du]{jin2024impact}
Mingyu Jin, Qinkai Yu, Dong Shu, Haiyan Zhao, Wenyue Hua, Yanda Meng, Yongfeng Zhang, and Mengnan Du.
\newblock The impact of reasoning step length on large language models.
\newblock In \emph{Findings of the Association for Computational Linguistics: ACL 2024}, pp.\  1830--1842, 2024.

\bibitem[Kang et~al.(2025)Kang, Sun, Chen, and Zou]{kang2025c3ot}
Yu~Kang, Xianghui Sun, Liangyu Chen, and Wei Zou.
\newblock C3ot: Generating shorter chain-of-thought without compromising effectiveness.
\newblock In \emph{Proceedings of the AAAI Conference on Artificial Intelligence}, volume~39, pp.\  24312--24320, 2025.

\bibitem[Kojima et~al.(2022)Kojima, Gu, Reid, Matsuo, and Iwasawa]{kojima2022large}
Takeshi Kojima, Shixiang~Shane Gu, Machel Reid, Yutaka Matsuo, and Yusuke Iwasawa.
\newblock Large language models are zero-shot reasoners.
\newblock \emph{Advances in neural information processing systems}, 35:\penalty0 22199--22213, 2022.

\bibitem[Lee et~al.(2025)Lee, Che, and Peng]{lee2025well}
Ayeong Lee, Ethan Che, and Tianyi Peng.
\newblock How well do llms compress their own chain-of-thought? a token complexity approach.
\newblock \emph{arXiv preprint arXiv:2503.01141}, 2025.

\bibitem[Li et~al.(2025)Li, Cao, Cao, Li, Tan, Keutzer, Xing, Gonzalez, and Stoica]{li2025s}
Dacheng Li, Shiyi Cao, Chengkun Cao, Xiuyu Li, Shangyin Tan, Kurt Keutzer, Jiarong Xing, Joseph~E Gonzalez, and Ion Stoica.
\newblock S*: Test time scaling for code generation.
\newblock In \emph{EMNLP (Findings)}, pp.\  15964--15978, 2025.

\bibitem[Li et~al.(2026)Li, Qin, Wang, Ma, Chen, Li, and Liang]{li2026stop}
Jiakai Li, Ke~Qin, Rongzheng Wang, Yizhuo Ma, Qizhi Chen, Muquan Li, and Shuang Liang.
\newblock Stop when further reasoning won't help: Attention-state adaptive generation in reasoning models.
\newblock \emph{arXiv preprint arXiv:2606.15070}, 2026.

\bibitem[Liang et~al.(2025)Liang, Zhong, Yang, and Quan]{liang2025thinkswitcher}
Guosheng Liang, Longguang Zhong, Ziyi Yang, and Xiaojun Quan.
\newblock Thinkswitcher: When to think hard, when to think fast.
\newblock In \emph{EMNLP (Findings)}, pp.\  5185--5201, 2025.

\bibitem[Liao et~al.(2025)Liao, Xu, Dong, Li, Monz, Savarese, Sahoo, and Xiong]{liao2025reward}
Baohao Liao, Yuhui Xu, Hanze Dong, Junnan Li, Christof Monz, Silvio Savarese, Doyen Sahoo, and Caiming Xiong.
\newblock Reward-guided speculative decoding for efficient llm reasoning.
\newblock \emph{arXiv preprint arXiv:2501.19324}, 2025.

\bibitem[Lieder \& Griffiths(2017)Lieder and Griffiths]{lieder2017strategy}
Falk Lieder and Thomas~L Griffiths.
\newblock Strategy selection as rational metareasoning.
\newblock \emph{Psychological review}, 124\penalty0 (6):\penalty0 762, 2017.

\bibitem[Liu et~al.(2026{\natexlab{a}})Liu, Hu, Chu, and Choi]{liu2026diffadapt}
Xiang Liu, Xuming Hu, Xiaowen Chu, and Eunsol Choi.
\newblock Diffadapt: Difficulty-adaptive reasoning for token-efficient llm inference.
\newblock In \emph{International Conference on Learning Representations}, volume 2026, pp.\  31640--31666, 2026{\natexlab{a}}.

\bibitem[Liu \& Wang(2025)Liu and Wang]{liu2025answer}
Xin Liu and Lu~Wang.
\newblock Answer convergence as a signal for early stopping in reasoning.
\newblock In \emph{Proceedings of the 2025 Conference on Empirical Methods in Natural Language Processing}, pp.\  17907--17918, 2025.

\bibitem[Liu et~al.(2026{\natexlab{b}})Liu, Li, Ma, Zhang, and Guo]{liu2026think}
Yongjiang Liu, Haoxi Li, Xiaosong Ma, Jie Zhang, and Song Guo.
\newblock Think how to think: Mitigating overthinking with autonomous difficulty cognition in large reasoning models.
\newblock In \emph{Proceedings of the 64th Annual Meeting of the Association for Computational Linguistics (Volume 1: Long Papers)}, pp.\  38105--38126, 2026{\natexlab{b}}.

\bibitem[Liu et~al.(2025)Liu, Chen, Li, Qi, Pang, Du, Lee, and Lin]{liu2025understanding}
Zichen Liu, Changyu Chen, Wenjun Li, Penghui Qi, Tianyu Pang, Chao Du, Wee~Sun Lee, and Min Lin.
\newblock Understanding r1-zero-like training: A critical perspective.
\newblock \emph{arXiv preprint arXiv:2503.20783}, 2025.

\bibitem[Luo et~al.(2026)Luo, He, Wang, Liu, Li, Cao, Tao, Tan, and Shen]{luo2026o1}
Haotian Luo, Haiying He, Yibo Wang, Shiwei Liu, Wei Li, Xiaochun Cao, Dacheng Tao, Naiqiang Tan, and Li~Shen.
\newblock O1-pruner: Length-harmonizing fine-tuning for o1-like reasoning pruning.
\newblock In \emph{Findings of the Association for Computational Linguistics: ACL 2026}, pp.\  14242--14257, 2026.

\bibitem[Ma et~al.(2025{\natexlab{a}})Ma, He, Snell, Griggs, Min, and Zaharia]{ma2025reasoning}
Wenjie Ma, Jingxuan He, Charlie Snell, Tyler Griggs, Sewon Min, and Matei Zaharia.
\newblock Reasoning models can be effective without thinking.
\newblock \emph{arXiv preprint arXiv:2504.09858}, 2025{\natexlab{a}}.

\bibitem[Ma et~al.(2025{\natexlab{b}})Ma, Wan, Yu, Fang, and Wang]{ma2025cot}
Xinyin Ma, Guangnian Wan, Runpeng Yu, Gongfan Fang, and Xinchao Wang.
\newblock Cot-valve: Length-compressible chain-of-thought tuning.
\newblock In \emph{Proceedings of the 63rd Annual Meeting of the Association for Computational Linguistics (Volume 1: Long Papers)}, pp.\  6025--6035, 2025{\natexlab{b}}.

\bibitem[{MAA Committees}()]{maaAIME}
{MAA Committees}.
\newblock {AIME} problems and solutions.
\newblock URL \url{https://artofproblemsolving.com/wiki/index.php/AIME_Problems_and_Solutions}.

\bibitem[Madaan et~al.(2023)Madaan, Tandon, Gupta, Hallinan, Gao, Wiegreffe, Alon, Dziri, Prabhumoye, Yang, et~al.]{madaan2023self}
Aman Madaan, Niket Tandon, Prakhar Gupta, Skyler Hallinan, Luyu Gao, Sarah Wiegreffe, Uri Alon, Nouha Dziri, Shrimai Prabhumoye, Yiming Yang, et~al.
\newblock Self-refine: Iterative refinement with self-feedback.
\newblock \emph{Advances in neural information processing systems}, 36:\penalty0 46534--46594, 2023.

\bibitem[Nelson(1990)]{nelson1990metamemory}
Thomas~O Nelson.
\newblock Metamemory: A theoretical framework and new findings.
\newblock In \emph{Psychology of learning and motivation}, volume~26, pp.\  125--173. Elsevier, 1990.

\bibitem[Pan et~al.(2025)Pan, Dai, Zhang, Oliaro, Jia, and Netravali]{pan2025specreason}
Rui Pan, Yinwei Dai, Zhihao Zhang, Gabriele Oliaro, Zhihao Jia, and Ravi Netravali.
\newblock Specreason: Fast and accurate inference-time compute via speculative reasoning.
\newblock In \emph{The Thirty-ninth Annual Conference on Neural Information Processing Systems}, 2025.
\newblock URL \url{https://openreview.net/forum?id=wCbOKbZ7kf}.

\bibitem[Phan et~al.(2025)Phan, Gatti, Han, Li, Hu, Zhang, Zhang, Shaaban, Ling, Shi, et~al.]{phan2025humanity}
Long Phan, Alice Gatti, Ziwen Han, Nathaniel Li, Josephina Hu, Hugh Zhang, Chen Bo~Calvin Zhang, Mohamed Shaaban, John Ling, Sean Shi, et~al.
\newblock Humanity's last exam.
\newblock \emph{arXiv preprint arXiv:2501.14249}, 2025.

\bibitem[{Qwen Team}(2026)]{qwen38}
{Qwen Team}.
\newblock Qwen3.8-max: A new bar for coding and cowork, August 2026.
\newblock URL \url{https://qwen.ai/blog?id=qwen3.8}.

\bibitem[Rein et~al.(2024)Rein, Hou, Stickland, Petty, Pang, Dirani, Michael, and Bowman]{rein2024gpqa}
David Rein, Betty~Li Hou, Asa~Cooper Stickland, Jackson Petty, Richard~Yuanzhe Pang, Julien Dirani, Julian Michael, and Samuel~R. Bowman.
\newblock {GPQA}: A graduate-level google-proof q\&a benchmark.
\newblock In \emph{First Conference on Language Modeling}, 2024.
\newblock URL \url{https://openreview.net/forum?id=Ti67584b98}.

\bibitem[Renze \& Guven(2024)Renze and Guven]{renze2024benefits}
Matthew Renze and Erhan Guven.
\newblock The benefits of a concise chain of thought on problem-solving in large language models.
\newblock \emph{arXiv preprint arXiv:2401.05618}, 2024.

\bibitem[Shao et~al.(2024)Shao, Wang, Zhu, Xu, Song, Bi, Zhang, Zhang, Li, Wu, et~al.]{shao2024deepseekmath}
Zhihong Shao, Peiyi Wang, Qihao Zhu, Runxin Xu, Junxiao Song, Xiao Bi, Haowei Zhang, Mingchuan Zhang, YK~Li, Yang Wu, et~al.
\newblock Deepseekmath: Pushing the limits of mathematical reasoning in open language models.
\newblock \emph{arXiv preprint arXiv:2402.03300}, 2024.

\bibitem[Shen et~al.(2025)Shen, Zhang, Huang, Shi, Zhang, Yan, Wang, Wang, Liu, and Lian]{shen2025dast}
Yi~Shen, Jian Zhang, Jieyun Huang, Shuming Shi, Wenjing Zhang, Jiangze Yan, Ning Wang, Kai Wang, Zhaoxiang Liu, and Shiguo Lian.
\newblock Dast: Difficulty-adaptive slow-thinking for large reasoning models.
\newblock In \emph{Proceedings of the 2025 Conference on Empirical Methods in Natural Language Processing: Industry Track}, pp.\  2322--2331, 2025.

\bibitem[Snell et~al.(2025)Snell, Lee, Xu, and Kumar]{snell2025scaling}
Charlie Snell, Jaehoon Lee, Kelvin Xu, and Aviral Kumar.
\newblock Scaling llm test-time compute optimally can be more effective than scaling parameters for reasoning.
\newblock In \emph{International Conference on Learning Representations}, volume 2025, pp.\  10131--10165, 2025.

\bibitem[Sprague et~al.(2025)Sprague, Yin, Rodriguez, Jiang, Wadhwa, Singhal, Zhao, Ye, Mahowald, and Durrett]{sprague2025cot}
Zayne Sprague, Fangcong Yin, Juan Rodriguez, Dongwei Jiang, Manya Wadhwa, Prasann Singhal, Xinyu Zhao, Xi~Ye, Kyle Mahowald, and Greg Durrett.
\newblock To cot or not to cot? chain-of-thought helps mainly on math and symbolic reasoning.
\newblock In \emph{International Conference on Learning Representations}, volume 2025, pp.\  94118--94162, 2025.

\bibitem[Sui et~al.(2025)Sui, Chuang, Wang, Zhang, Zhang, Yuan, Liu, Wen, Zhong, Zou, et~al.]{sui2025stop}
Yang Sui, Yu-Neng Chuang, Guanchu Wang, Jiamu Zhang, Tianyi Zhang, Jiayi Yuan, Hongyi Liu, Andrew Wen, Shaochen Zhong, Na~Zou, et~al.
\newblock Stop overthinking: A survey on efficient reasoning for large language models.
\newblock \emph{arXiv preprint arXiv:2503.16419}, 2025.

\bibitem[Tikhonov et~al.(2026)Tikhonov, Oseledets, and Tutubalina]{tikhonov2026confidence}
Pavel Tikhonov, Ivan Oseledets, and Elena Tutubalina.
\newblock Confidence leaps in llm reasoning: Early stopping and cross-model transfer.
\newblock In \emph{Proceedings of the 19th Conference of the European Chapter of the Association for Computational Linguistics (Volume 2: Short Papers)}, pp.\  602--616, 2026.

\bibitem[Wang et~al.(2026)Wang, Hu, Chen, Wang, Chen, Lin, Rong, Li, Wen, and Tan]{wang2026intervene}
Qianyue Wang, Jinwu Hu, Yaofo Chen, Yufeng Wang, Bailin Chen, Huanxiang Lin, Yu~Rong, Yuanqing Li, Zhiquan Wen, and Mingkui Tan.
\newblock Intervene when it doubts: Conjunction-guided interactive reasoning.
\newblock In \emph{Forty-third International Conference on Machine Learning}, 2026.
\newblock URL \url{https://openreview.net/forum?id=zVjLO7jg9T}.

\bibitem[Wang et~al.(2025)Wang, Zhang, Huang, Yang, Zhang, Huang, and Wang]{wang2025sampling}
Yiming Wang, Pei Zhang, Siyuan Huang, Baosong Yang, Zhuosheng Zhang, Fei Huang, and Rui Wang.
\newblock Sampling-efficient test-time scaling: Self-estimating the best-of-n sampling in early decoding.
\newblock In \emph{The Thirty-ninth Annual Conference on Neural Information Processing Systems}, 2025.
\newblock URL \url{https://openreview.net/forum?id=BcKYVmh3yH}.

\bibitem[Wei et~al.(2022)Wei, Wang, Schuurmans, Bosma, Xia, Chi, Le, Zhou, et~al.]{wei2022chain}
Jason Wei, Xuezhi Wang, Dale Schuurmans, Maarten Bosma, Fei Xia, Ed~Chi, Quoc~V Le, Denny Zhou, et~al.
\newblock Chain-of-thought prompting elicits reasoning in large language models.
\newblock \emph{Advances in neural information processing systems}, 35:\penalty0 24824--24837, 2022.

\bibitem[Xia et~al.(2025)Xia, Leong, Wang, Li, and Li]{xia2025tokenskip}
Heming Xia, Chak~Tou Leong, Wenjie Wang, Yongqi Li, and Wenjie Li.
\newblock Tokenskip: Controllable chain-of-thought compression in llms.
\newblock In \emph{Proceedings of the 2025 Conference on Empirical Methods in Natural Language Processing}, pp.\  3351--3363, 2025.

\bibitem[Xia et~al.(2026)Xia, Xie, Xu, Kang, Lamba, Gao, and McAuley]{xia2026agentic}
Yu~Xia, Zhouhang Xie, Xin Xu, Byungkyu Kang, Prarit Lamba, Xiang Gao, and Julian McAuley.
\newblock Agentic chain-of-thought steering for efficient and controllable llm reasoning.
\newblock \emph{arXiv preprint arXiv:2606.03965}, 2026.

\bibitem[Xu et~al.(2026)Xu, Lin, Xue, Wang, Xu, Wu, Zhang, Lin, Dong, Ling, et~al.]{xu2026deepseek}
Anyi Xu, Bangcai Lin, Bing Xue, Bingxuan Wang, Bingzheng Xu, Bochao Wu, Bowei Zhang, Chaofan Lin, Chen Dong, Chenchen Ling, et~al.
\newblock Deepseek-v4: Towards highly efficient million-token context intelligence.
\newblock \emph{arXiv preprint arXiv:2606.19348}, 2026.

\bibitem[Xu et~al.(2025)Xu, Xie, Zhao, and He]{xu2025chain}
Silei Xu, Wenhao Xie, Lingxiao Zhao, and Pengcheng He.
\newblock Chain of draft: Thinking faster by writing less.
\newblock \emph{arXiv preprint arXiv:2502.18600}, 2025.

\bibitem[Yang et~al.(2025{\natexlab{a}})Yang, Li, Yang, Zhang, Hui, Zheng, Yu, Gao, Huang, Lv, et~al.]{yang2025qwen3}
An~Yang, Anfeng Li, Baosong Yang, Beichen Zhang, Binyuan Hui, Bo~Zheng, Bowen Yu, Chang Gao, Chengen Huang, Chenxu Lv, et~al.
\newblock Qwen3 technical report.
\newblock \emph{arXiv preprint arXiv:2505.09388}, 2025{\natexlab{a}}.

\bibitem[Yang et~al.(2026)Yang, Si, Duan, Zhu, Zhu, Li, Chen, Lin, and Wang]{yang2026dynamic}
Chenxu Yang, Qingyi Si, Yongjie Duan, Zheliang Zhu, Chenyu Zhu, Qiaowei Li, Minghui Chen, Zheng Lin, and Weipinng Wang.
\newblock Dynamic early exit in reasoning models.
\newblock In \emph{International Conference on Learning Representations}, volume 2026, pp.\  88170--88210, 2026.

\bibitem[Yang et~al.(2025{\natexlab{b}})Yang, Lin, and Yu]{yang2025think}
Junjie Yang, Ke~Lin, and Xing Yu.
\newblock Think when you need: Self-adaptive chain-of-thought learning.
\newblock \emph{arXiv preprint arXiv:2504.03234}, 2025{\natexlab{b}}.

\bibitem[Yang et~al.(2025{\natexlab{c}})Yang, Yue, Chaudhary, and Han]{yang2025speculative}
Wang Yang, Xiang Yue, Vipin Chaudhary, and Xiaotian Han.
\newblock Speculative thinking: Enhancing small-model reasoning with large model guidance at inference time.
\newblock \emph{arXiv preprint arXiv:2504.12329}, 2025{\natexlab{c}}.

\bibitem[Yang et~al.(2025{\natexlab{d}})Yang, Ma, Lin, and Wei]{yang2025towards}
Wenkai Yang, Shuming Ma, Yankai Lin, and Furu Wei.
\newblock Towards thinking-optimal scaling of test-time compute for {LLM} reasoning.
\newblock In \emph{The Thirty-ninth Annual Conference on Neural Information Processing Systems}, 2025{\natexlab{d}}.
\newblock URL \url{https://openreview.net/forum?id=6ICFqmixlS}.

\bibitem[Yao et~al.(2023)Yao, Yu, Zhao, Shafran, Griffiths, Cao, and Narasimhan]{yao2023tree}
Shunyu Yao, Dian Yu, Jeffrey Zhao, Izhak Shafran, Tom Griffiths, Yuan Cao, and Karthik Narasimhan.
\newblock Tree of thoughts: Deliberate problem solving with large language models.
\newblock \emph{Advances in neural information processing systems}, 36:\penalty0 11809--11822, 2023.

\bibitem[Yu et~al.(2024)Yu, Xu, Weston, and Kulikov]{yu2024distilling}
Ping Yu, Jing Xu, Jason Weston, and Ilia Kulikov.
\newblock Distilling system 2 into system 1.
\newblock \emph{arXiv preprint arXiv:2407.06023}, 2024.

\bibitem[Yu et~al.(2025{\natexlab{a}})Yu, Yu, and Wang]{yu2025premise}
Ye~Yu, Yaoning Yu, and Haohan Wang.
\newblock Premise: Scalable and strategic prompt optimization for efficient mathematical reasoning in large models.
\newblock \emph{arXiv preprint arXiv:2506.10716}, 2025{\natexlab{a}}.

\bibitem[Yu et~al.(2025{\natexlab{b}})Yu, Zhang, Zhang, Liang, Zhang, Zhang, Khademi, Awadalla, Wang, Yang, et~al.]{yu2025chain}
Yiyao Yu, Yuxiang Zhang, Dongdong Zhang, Xiao Liang, Hengyuan Zhang, Xingxing Zhang, Mahmoud Khademi, Hany~Hassan Awadalla, Junjie Wang, Yujiu Yang, et~al.
\newblock Chain-of-reasoning: Towards unified mathematical reasoning in large language models via a multi-paradigm perspective.
\newblock In \emph{Proceedings of the 63rd Annual Meeting of the Association for Computational Linguistics (Volume 1: Long Papers)}, pp.\  24914--24937, 2025{\natexlab{b}}.

\bibitem[Zeng et~al.(2026)Zeng, Lv, Hou, Du, Zheng, Chen, Yin, Ge, Huang, Xie, et~al.]{zeng2026glm}
Aohan Zeng, Xin Lv, Zhenyu Hou, Zhengxiao Du, Qinkai Zheng, Bin Chen, Da~Yin, Chendi Ge, Chenghua Huang, Chengxing Xie, et~al.
\newblock Glm-5: from vibe coding to agentic engineering.
\newblock \emph{arXiv preprint arXiv:2602.15763}, 2026.

\bibitem[Zhang et~al.(2025)Zhang, Xiao, and Cao]{zhang2025long}
Ruiqi Zhang, Changyi Xiao, and Yixin Cao.
\newblock Long or short cot? investigating instance-level switch of large reasoning models.
\newblock \emph{arXiv preprint arXiv:2506.04182}, 2025.

\bibitem[Zhang et~al.(2026)Zhang, Mohananey, Chronopoulou, Papalampidi, Gupta, Munkhdalai, Wang, and Upadhyay]{zhang2026llms}
Xinliang~Frederick Zhang, Anhad Mohananey, Alexandra Chronopoulou, Pinelopi Papalampidi, Somit Gupta, Tsendsuren Munkhdalai, Lu~Wang, and Shyam Upadhyay.
\newblock Do llms really need 10+ thoughts for “find the time 1000 days later”? towards structural understanding of llm overthinking.
\newblock In \emph{Proceedings of the 64th Annual Meeting of the Association for Computational Linguistics (Volume 1: Long Papers)}, pp.\  17005--17030, 2026.

\bibitem[Zheng et~al.(2024)Zheng, Yin, Xie, Sun, Huang, Yu, Cao, Kozyrakis, Stoica, Gonzalez, et~al.]{zheng2024sglang}
Lianmin Zheng, Liangsheng Yin, Zhiqiang Xie, Chuyue Sun, Jeff Huang, Cody~H Yu, Shiyi Cao, Christos Kozyrakis, Ion Stoica, Joseph~E Gonzalez, et~al.
\newblock Sglang: Efficient execution of structured language model programs.
\newblock \emph{Advances in neural information processing systems}, 37:\penalty0 62557--62583, 2024.

\bibitem[Zhou et~al.(2026)Zhou, Ling, Chen, Wang, Fan, and Wang]{zhou2026more}
Shu Zhou, Rui Ling, Junan Chen, Xin Wang, Tao Fan, and Hao Wang.
\newblock When more thinking hurts: Overthinking in llm test-time compute scaling.
\newblock In \emph{Findings of the Association for Computational Linguistics: ACL 2026}, pp.\  23967--23977, 2026.

\bibitem[Zhu et~al.(2025)Zhu, Xie, Lv, and slime Contributors]{slime_github}
Zilin Zhu, Chengxing Xie, Xin Lv, and slime Contributors.
\newblock slime: An llm post-training framework for rl scaling.
\newblock \url{https://github.com/THUDM/slime}, 2025.
\newblock GitHub repository. Corresponding author: Xin Lv.

\end{thebibliography}
